\documentclass{article}

\usepackage{microtype}
\usepackage{graphicx}
\usepackage{booktabs} 

\usepackage{hyperref}

\usepackage[accepted]{icml2024}

\usepackage{amsmath}
\usepackage{amssymb}
\usepackage{mathtools}
\usepackage{amsthm}

\usepackage[capitalize,noabbrev]{cleveref}

\newcommand{\tabincell}[2]{\begin{tabular}{@{}#1@{}}#2\end{tabular}}
\usepackage{boldline}
\usepackage{colortbl}
\usepackage{graphicx}
\definecolor{mygray}{gray}{.9}
\usepackage{algorithm}
\usepackage{algorithmic}
\usepackage{color}
\usepackage{multirow}
\usepackage{amsfonts,amssymb}
\usepackage{amsmath}
\usepackage{booktabs}
\usepackage{mathtools}
\usepackage{url}
\usepackage{stfloats}
\usepackage{bbm}
\usepackage{float}
\usepackage[caption=false,font=footnotesize,farskip=0pt]{subfig}
\usepackage{amsthm}
\usepackage{enumerate}

\usepackage{enumitem}
\definecolor{MyGreen}{RGB}{6,128,67}
\definecolor{HLcolor}{RGB}{164,218,246}
\usepackage{makecell}
\usepackage{supertabular}
\usepackage{amsmath}
\usepackage{boldline}
\usepackage{soul}

\usepackage{graphicx}

\usepackage[textsize=tiny]{todonotes}

\icmltitlerunning{Are Large Language Models Good Prompt Optimizers?}

\begin{document}

\twocolumn[
\icmltitle{Are Large Language Models Good Prompt Optimizers?}



\icmlsetsymbol{equal}{*}
\icmlsetsymbol{intern}{\dag}
\icmlsetsymbol{correspond}{\ddag}

\begin{icmlauthorlist}
\icmlauthor{Ruotian Ma}{yyy,comp,equal,intern}
\icmlauthor{Xiaolei Wang}{yyy,equal}
\icmlauthor{Xin Zhou}{yyy}
\icmlauthor{Jian Li}{comp}
\icmlauthor{Nan Du}{comp,correspond}
\icmlauthor{Tao Gui}{sch}
\icmlauthor{Qi Zhang}{yyy,correspond}
\icmlauthor{Xuanjing Huang}{yyy}
\end{icmlauthorlist}

\center{$^1$ School of Computer Science, Fudan University\\
\ \ \ $^2$ Tencent AI Lab \ \ \ \ \  \ $^3$ Institute of Modern Languages and Linguistics, Fudan University\\
\tt
rtma19@fudan.edu.cn, xlwang22@m.fudan.edu.cn}


\icmlaffiliation{yyy}{School of Computer Science, Fudan University, Shanghai, China}
\icmlaffiliation{comp}{Tencent AI Lab}
\icmlaffiliation{sch}{Institute of Modern Languages and Linguistics, Fudan University, Shanghai, China}

\icmlcorrespondingauthor{Ruotian Ma}{rtma19@fudan.edu.cn}
\icmlcorrespondingauthor{Nan Du}{frankndu@tencent.com}
\icmlcorrespondingauthor{Qi Zhang}{qz@fudan.edu.cn}

\icmlkeywords{Machine Learning, ICML}

\vskip 0.3in
]



\printAffiliationsAndNotice{\icmlEqualContribution} 

\begin{abstract}
LLM-based Automatic Prompt Optimization, which typically utilizes LLMs as Prompt Optimizers to self-reflect and refine prompts, has shown promising performance in recent studies. Despite the success, the underlying mechanism of this approach remains unexplored, and the true effectiveness of LLMs as Prompt Optimizers requires further validation. In this work, we conducted a comprehensive study to uncover the actual mechanism of LLM-based Prompt Optimization. Our findings reveal 
that the LLM optimizers struggle to identify the true causes of errors during reflection, tending to be biased by their own prior knowledge rather than genuinely reflecting on the errors. Furthermore, even when the reflection is semantically valid, the LLM optimizers often fail to generate appropriate prompts for the target models with a single prompt refinement step, partly due to the unpredictable behaviors of the target models. Based on the observations, we introduce a new ``Automatic Behavior Optimization'' paradigm, which directly optimizes the target model's behavior in a more controllable manner. We hope our study can inspire new directions for automatic prompt optimization development.
\end{abstract}

\section{Introduction}

Prompt Engineering, the practice of designing optimal prompts for maximizing the performance of large language models (LLMs), has become increasingly significant in the current landscape of LLM applications \cite{brown2020language,gao-etal-2021-making,wei2022chain, wei2021finetuned, wang2022self, liu2023pre}. Crafting suitable prompts often demands significant human effort, specialized expertise, and extensive trial-and-error iterations \cite{zamfirescu2023johnny}. Hence, exploring effective automatic prompt engineering methods becomes crucial for reducing costs \cite{shin2020autoprompt, deng2022rlprompt, zhang2022tempera, prasad2022grips,chen2023instructzero}.

Recently, LLM-based Automatic Prompt Optimization has been widely explored \cite{zhou2023large,pryzant2023automatic}. These works typically utilize LLMs as prompt optimizers, to iteratively refine prompts for target models. These methods include Resampling-based methods, which employ LLMs to generate semantically similar prompt variations \cite{zhou2023large, li-etal-2023-robust}, and Reflection-based methods, where LLMs optimizers optimize prompts through self-reflection on errors or historical prompts \cite{pryzant2023automatic, sun2023autohint, ye2023prompt,wang2023promptagent,yang2023large,guo2023connecting}. Particularly in Reflection-based methods, LLM-based prompt optimizers act like expert prompt engineers by mirroring the trial-and-error process in human-level prompt design. As a result, the paradigm of LLMs as Prompt Optimizers has achieved promising advancements and widespread interest.

Despite the success of LLMs as Prompt Optimizers, the underlying mechanism of the LLM-based Automatic Prompt Optimization process remains underexplored. As recent research has highlighted the limitations of LLMs on self-correction \cite{stechly2023gpt, valmeekam2023can,huang2023large,tyen2023llms}, it also cast doubt on the true proficiency of LLMs in reflecting and refining prompts. This prompts us to conduct a comprehensive study to critically assess the effectiveness of LLMs as prompt optimizers.

Firstly, we are to validate the actual effectiveness of LLMs as prompt optimizers. Recent research has revealed that, aside from the designs of the prompt optimizers, variations in search strategy and prompt initialization designs also largely affect LLM-based prompt optimization performance \cite{ye2023prompt,zhang-etal-2023-auto}. Thus, it is crucial to isolate the effect of the LLM prompt optimizers by standardizing all other factors. In this study, we standardize the implementations of various methods with a unified setting, to examine the true effectiveness of LLMs as prompt optimizers. Our findings reveal that the Reflection-based prompt optimization methods do not consistently outperform the Resampling-based method. Additionally, sophisticated designs of the reflection and refinement process do not necessarily yield superior performance. These results indicate that the underlying mechanism of LLM-based prompt optimization may differ from our expectations.

We next conducted comprehensive experiments to investigate the true mechanism of LLMs as prompt optimizers from various views. Our primary findings can be summarized as follows:
\begin{itemize}[leftmargin=0.5em, itemindent=-0.1em, itemsep=0em]
    \item \textbf{Repetitive Reflection} (\S \ref{sec:reflection}): By thoroughly examining the reflection process of the LLM optimizers, we observed that the LLM optimizers consistently generate similar feedbacks regardless of the error distributions. Additionally, the feedbacks generated are significantly repetitive even across different steps. These results indicate that the LLM optimizers might struggle to uncover the true causes of errors. During reflection, the LLM optimizers may be biased by their own prior knowledge of the tasks, rather than genuinely reflecting on the errors.

    \item \textbf{Inappropriate Prompt Refinement} (\S\ref{sec:refinement}): As we delve further into the prompt refinement process, we verify that the Reflection-based prompt optimizers is able to make valid \textit{semantic alterations} on prompts in \textit{certain steps}. However, the LLM optimizers generally fail to generate appropriate prompts (for target models) within the altered semantic space through a single refinement process. We further show that the challenges faced by LLM optimizers in refining prompts can be partly attributed to the uncontrollable behaviors of the target models in following the instructions.
    
    \item \textbf{Advocating for New Paradigm}  (\S \ref{sec:behavior}): Our observations reveal a gap between the LLM optimizer and the target models in the existing LLM-based prompt optimization paradigm—the LLM optimizer struggles to understand the failures of the target models, while the target models face difficulties in appropriately following the generated instructions. Therefore, we advocate for exploring new LLM-based prompt optimization paradigms that can substantially alleviate these problems. As a preliminary attempt, we introduce an ``Automatic Behavior Optimization'' paradigm, which allows more directly and objectively refining target models' behaviors. We observed that Automatic Behavior Optimization is rather effective in improving the behaviors of less powerful target models. 
\end{itemize}

We hope our study can inspire the development of new paradigms and further advancements in the field of LLM-based Automatic Prompt Optimization.


\section{Background: LLM-based Automatic Prompt Optimization}

\begin{table*}[h]
\centering
\renewcommand\arraystretch{1.6}
\small
\caption{Various designs of existing LLM-based Prompt Optimization methods and the Unified Setting adopted in this work. We standardized the initialization and search strategy across different methods to achieve \textbf{a fair evaluation of various prompt updating methods}. ``$-$'' means the setting is not applicable to the method (in the case of PromptAgent, the search size per step is associated with the real-time process of Monte Carlo Tree Search). ``Unk.'' denotes settings not reported in the paper. ``$P_{t-1}$'' denotes the set of the prompts to be updated at each step $t$. }
\vspace{0.3cm}
\scalebox{0.75}{
\begin{tabular}{|l|c|c|c|c|c|c|c|c|c|c|c|}
\hline
\multirow{2}*{\textbf{\makecell[l]{\\Methods}}} &
\multicolumn{2}{c|}{\bfseries Prompt Initialization} & 
\multicolumn{5}{c|}{\bfseries Search Strategy}& 
\makecell[c]{\bfseries Prompt Updating}\\
\cline{2-9}
&{\footnotesize \makecell[c]{Method \\ Type}} 
& \makecell[c]{\footnotesize Attribute Involved \\ in Initialization} 
& \makecell[c]{\footnotesize Search \\ Algorithm} 
& {\footnotesize \makecell*[c]{Initial \\ size}} 
& \makecell[c]{\footnotesize Expansion \\ size per step} 
& \makecell[c]{\footnotesize Selection \\ size per step} 
& \makecell[c]{\footnotesize Total \\ Steps} & Method Type \\
\hline

\textbf{  Iterative-APE }\cite{zhou2023large} & LLM & \footnotesize Examples & Beam Search & $50$ & Unk. & Unk. & Unk. & Resampling \\

\textbf{  APO }\cite{pryzant2023automatic}  & Manual   &  Manual instruction  & Beam Search & $1$ & $|{P}_{t-1}|\times12$& $4$& $6$ & Explicit Reflection\\

\textbf{ APO-Sum }(this work)  & LLM  & Manual instruction$+$Examples & Beam Search &  $10$ & $|{P}_{t-1}|\times2$& $5$ & $10$ &Explicit Reflection\\

\textbf{  APO-Agent }\cite{wang2023promptagent} & Manual & Manual instruction  & \makecell[c]{\fontsize{8pt}{\baselineskip}\selectfont Monte Carlo \\ \fontsize{8pt}{\baselineskip}\selectfont Tree Search}   & $1$ & $-$ & $-$ & $12$ &Explicit Reflection\\

\textbf{  OPRO }\cite{yang2023large} & Manual  & Initialize with empty string  &$-$& $1$ & $8$ &$-$& $200$&Implicit Reflection\\

\textbf{ Unified Setting } (this work) &LLM  & Manual instruction$+$Examples  & Beam Search &  $10$ & $|{P}_{t-1}|\times2$& $5$ & $10$& $-$\\

\hline

\end{tabular}}
\label{tab:unifiy_design}
\vspace{-0.2cm}
\end{table*}

In this work, we focus on LLM-based Automatic Prompt Optimization, which leverages LLMs as prompt optimizers to obtain suitable prompts within discrete natural language spaces. Formally, consider a training set $\mathcal{D}_{train}= \{(x_i, y_i)\}_{i=1}^n$ drawn from a task $\mathcal{T}$ and a score function $s(\cdot)$ for the task. We are to perform the task with a target model $\mathcal{M}$, typically a black-box LLM. The goal of prompt optimization is to find the optimal prompt $p^*$ drawn from the natural language space that maximizes the expectation of the score over $\mathcal{D}_{train}$:
\begin{equation}
    p^* = \arg \max_p \mathbb{E}_{(x_i, y_i)\sim \mathcal{D}_{train}}[s(\mathcal{M}, p, x_i, y_i)]
\end{equation}

An LLM-based Automatic Prompt Optimization framework typically consists of three components (excluding the design of scorer $s$): prompt initialization, prompt updating, and search strategy. 

\paragraph{Prompt Initialization}
A prompt initialization process is to obtain a set of initial prompts ${P}_0=\{p_0^i\}_{i=1}^N$. Existing works have explored two typical prompt initialization ways: (1) Manual Prompt Initialization: using human-written task descriptions as initial prompts \cite{yang2023large, pryzant2023automatic}; (2) LLM-based Prompt Initialization: these methods leverage LLMs to generate initial prompts based on few-shot examples \cite{zhou2023large,zhang-etal-2023-auto}. 

\paragraph{Search Strategy}\label{sec:search}
The search strategy determines the prompt filter, selection, and prompt updating methods adopted in Automatic Prompt Optimization.
Typically, in each step, given a set of newly generated prompts $\hat{{P}}_{t}=\{p_{t-1}^i\}_{i=1}^N$ and the corresponding scores ${R}_{t}=\{s(\hat{p}_{t}^i)\}_{i=1}^N$, the search strategy decides a selection function $\mathcal{S}(\cdot)$ that determines the set of prompts $P_t = \mathcal{S}(\hat{P}_t,{R}_{t})$ to be selected for the next step. It also decides the number of prompts to expand in each step and the total search size.
Recent research illustrates the important role of the search strategy in affecting the overall prompt optimization performance \cite{wang2023promptagent,ye2023prompt}.

\paragraph{Prompt Updating}
The prompt updating module generates a new set of prompts $\hat{P}_t$ based on the prompts ${P}_{t-1}$ from the last step. 
In an LLM-based Prompt Optimization Framework, this process is fulfilled with an LLM-based prompt generator $\mathcal{G}$ or prompt optimizer $\mathcal{O}$.
Existing prompt updating methods can be categorized into three types:
\begin{itemize}[leftmargin=0.5em, itemindent=0em, itemsep=0em]
   \item Resampling-based Prompt Regeneration \cite{zhou2023large}: The core idea of this method is to sample around the current best prompts for better prompt candidates while keeping the semantic meanings. Typically, a prompt generator $\mathcal{G}$ receives the current prompt $p_{t-1}$ and generates variations of the current prompts without using any other information $\hat{p}_t = \mathcal{G}(p_{t-1})$. Here, we consider the role of the LLM prompt generator in resampling-based methods as a prompt regenerator rather than a prompt optimizer, since the resampling process is directionless without combined with the search strategy. We regard the resampling-based method as an important baseline for evaluating the effectiveness of the LLM-based prompt optimizers.

    \item Explicit Reflection-based Prompt Optimization \cite{pryzant2023automatic}: These methods explicitly leverage the self-reflection ability of LLMs for prompt refinement. Typically, given a current prompt $p_{t-1}$, an LLM-based prompt optimizer $\mathcal{O}$ analyzes on the errors and generates a reflection or feedback $c_{t-1} = \mathcal{O}(p_{t-1}, {E}_{t-1})$ regarding the current prompt $p_{t-1}$. Here, ${E}_{t-1}$ is a set of sampled error examples obtained from $\mathcal{M}$ with $p_{t-1}$. Next, the prompt optimizer refines the current prompt using the generated feedback by $\hat{p}_t = \mathcal{O}(p_{t-1}, c_{t-1}, {E}_{t-1})$. The practice of explicit reflection and refinement directly mirrors the trial-and-error process used in human prompt engineering. It also provides an interpretable insight into the behavior of LLM-based prompt optimizers. Therefore, it has become the most widely-explored method in recent studies \cite{pryzant2023automatic,sun2023autohint, ye2023prompt, wang2023promptagent,cheng2023blackbox}.

    \item Implicit Reflection-based Prompt Optimization \cite{yang2023large}: These method also employ an LLM prompt optimizer to generate new prompts based on historical prompts, scores, or error examples. Typically, an LLM prompt optimizer $O$ takes a set of historical prompts ${P}_{t-1}$ and their corresponding scores ${R}_{t-1}$ as input and directly generate a set of refined prompts $\hat{{P}}_{t}$ by $\hat{{P}}_{t}=\mathcal{O}({P}_{t-1}, \mathcal{R}_{t-1}, {E}_{t-1})$ (where ${E}_{t-1}$ is optional). Unlike Explicit Reflection-based Optimization, these methods do not explicitly generate any reflections during prompt refinement. However, they require the LLM to implicitly analyze the scores and errors related to the current prompts for prompt optimization, often guided by the meta prompt. Therefore, we refer to this practice as Implicit Reflection-based Optimization.
    
\end{itemize}

\section{Assessing LLMs as Prompt Optimizers}\label{sec:main_evaluation}

\begin{figure*}

\centering
\subfloat{\includegraphics[width=0.25\linewidth]{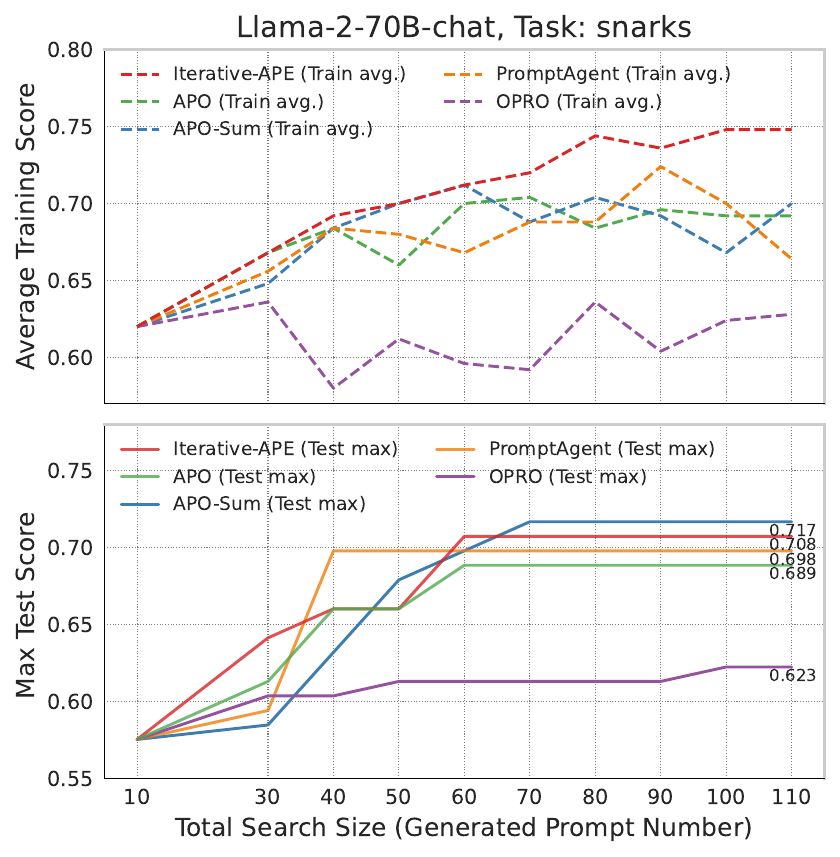}}
\subfloat{\includegraphics[width=0.25\linewidth]{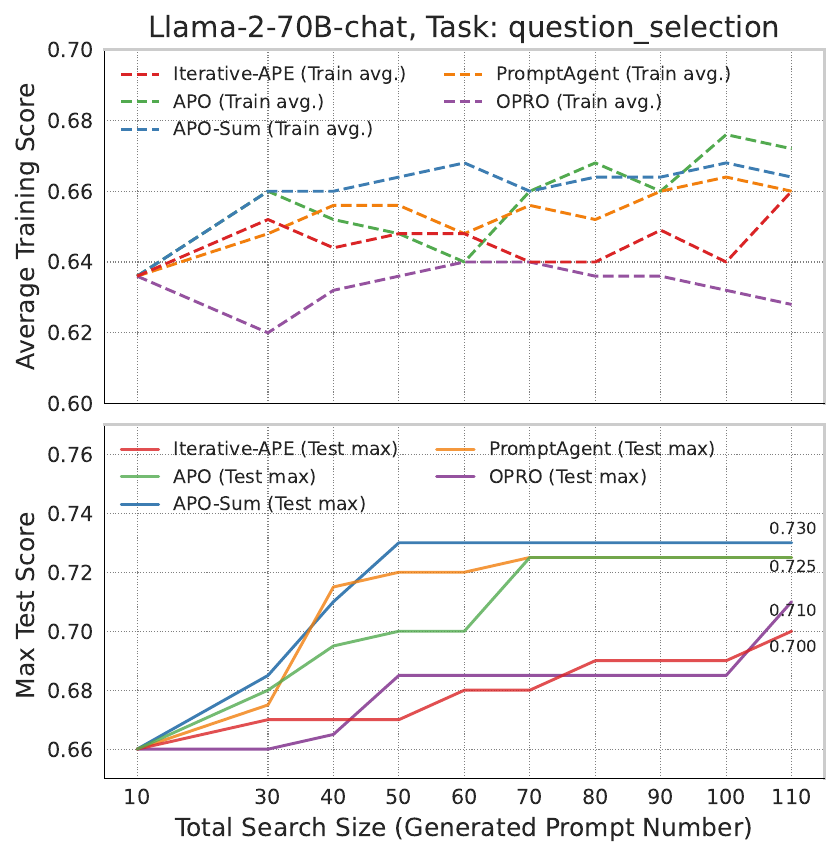}}
\subfloat{\includegraphics[width=0.25\linewidth]{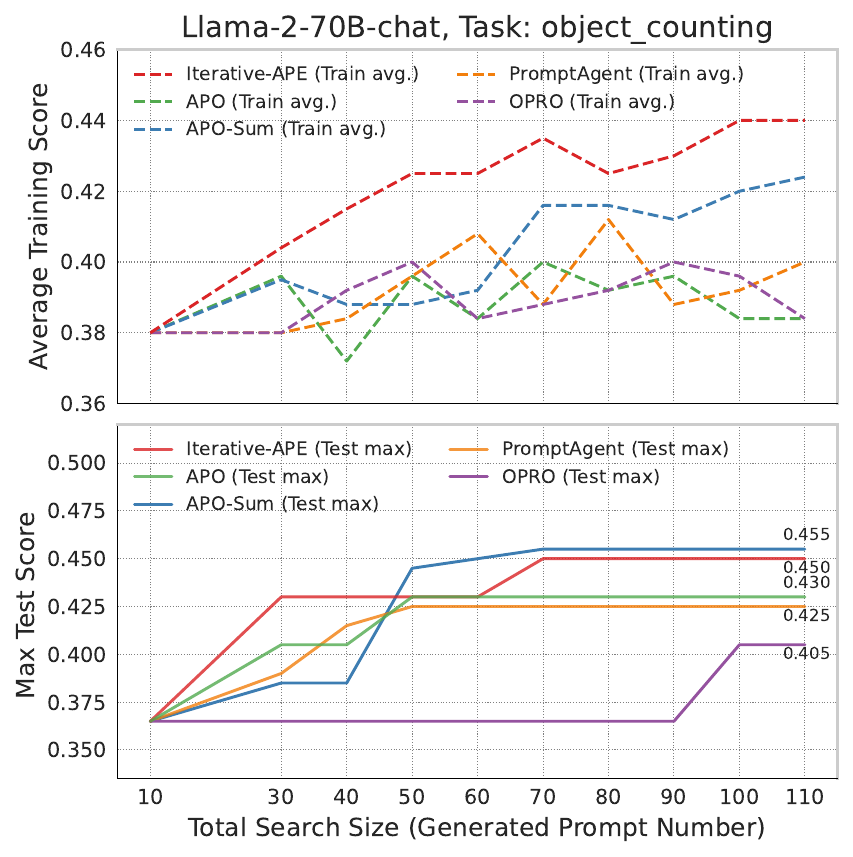}}
\subfloat{\includegraphics[width=0.25\linewidth]{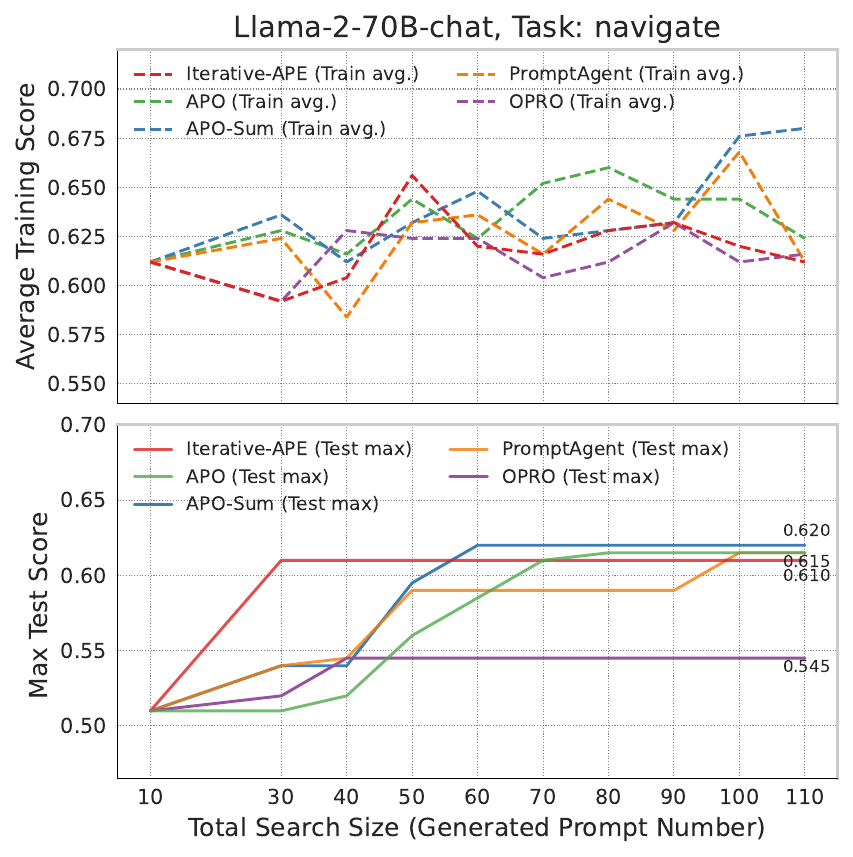}}\\
\subfloat{\includegraphics[width=0.25\linewidth]{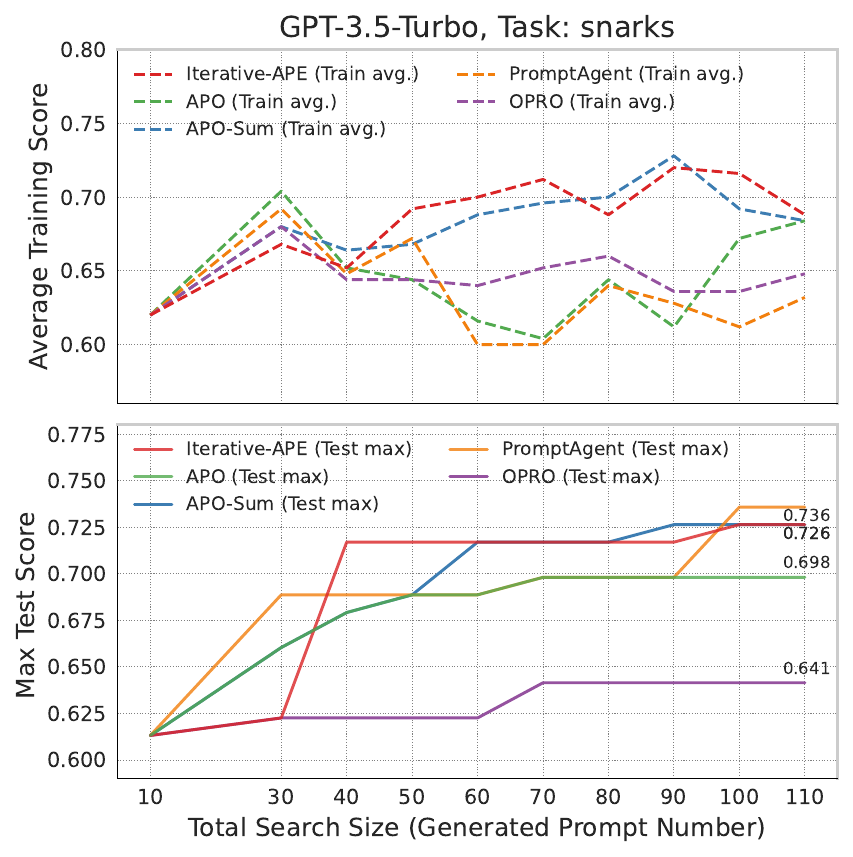}}
\subfloat{\includegraphics[width=0.25\linewidth]{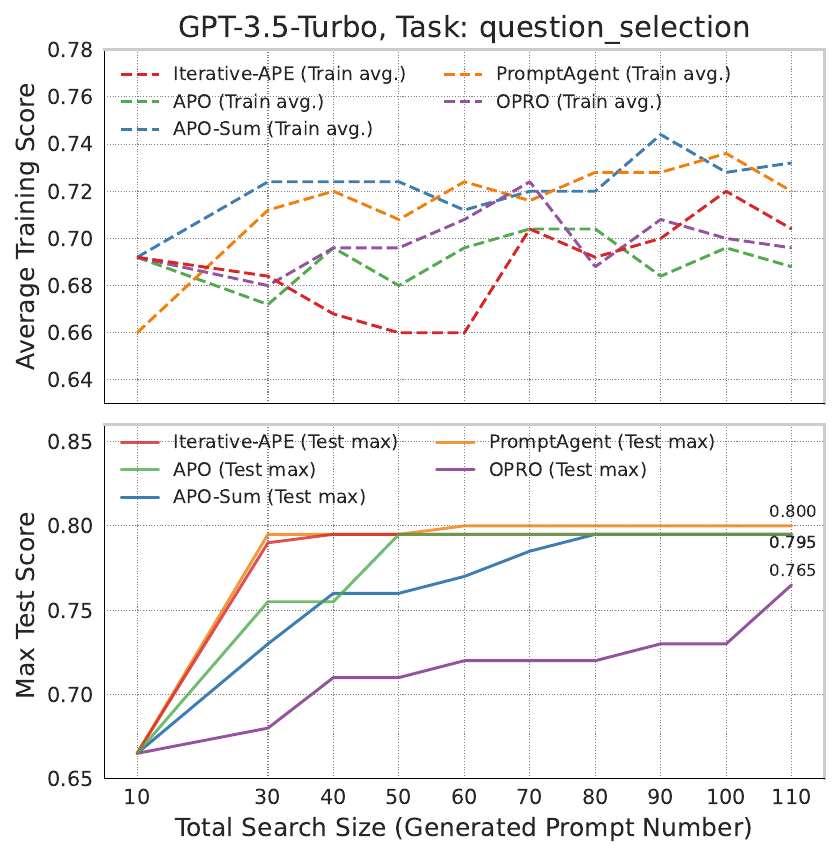}}
\subfloat{\includegraphics[width=0.25\linewidth]{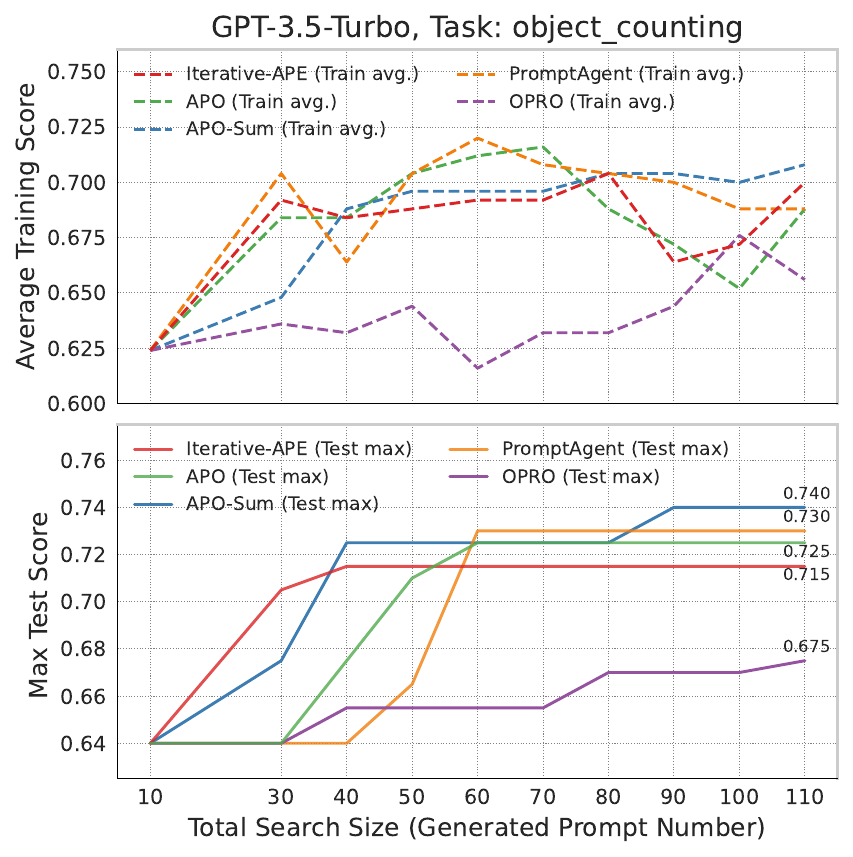}}
\subfloat{\includegraphics[width=0.25\linewidth]{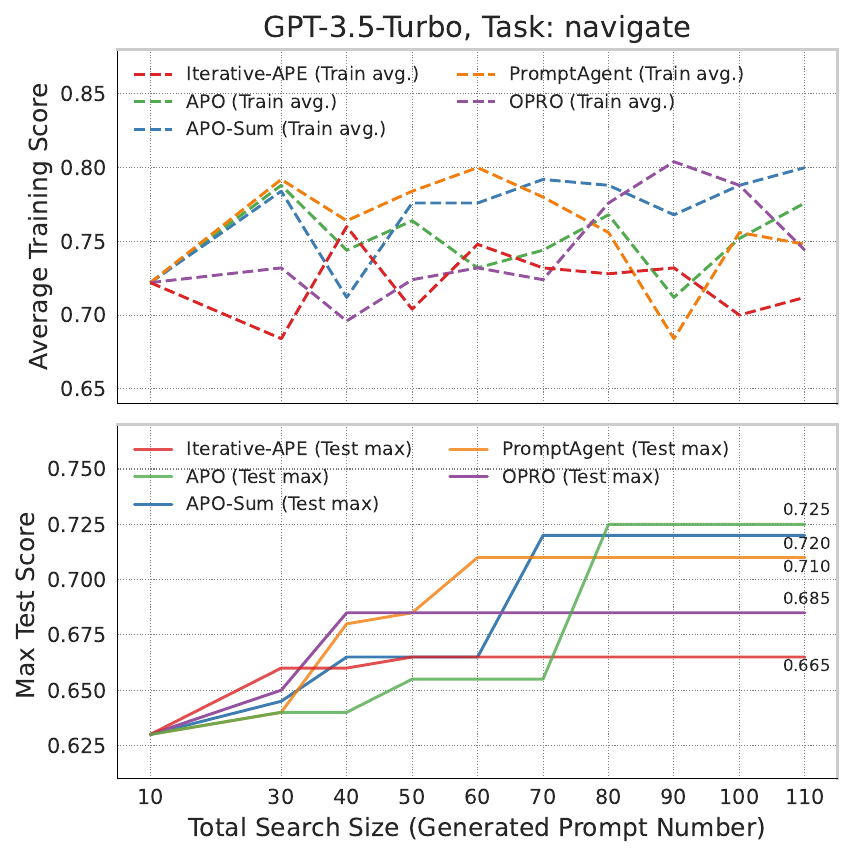}}

    \caption{ Results of various methods under the unified setting. We ran three repeated experiments and selected the best trend for each method. ``Max Test Score" denotes the highest accuracy of generated prompts to the current step.}
    \label{fig:results_unifying}
    \vspace{-0.3cm}
\end{figure*}

\subsection{Unifying the Implementation Designs}
In this work, we aim to investigate the LLM's behavior as prompt optimizers. Therefore, it is crucial to isolate and evaluate the impact of the LLM prompt optimizers within the entire prompt optimization framework. This necessitates excluding the impact of various prompt initialization and search strategies. To this end, we propose a Unified Setting for the LLM-based Prompt Optimization Framework. Detailed implementation designs are presented in Table \ref{tab:unifiy_design}.

\subsection{Experimental Settings} We compare the prompt updating process of five representative methods:

    \noindent \textbf{Iterative-APE} \cite{zhou2023large}: A typical resampling-based prompt regeneration method. In each step, the LLM is asked to generate variations of the current prompts while keeping their semantic meanings. As declared before, we consider the role of LLMs in Iterative-APE as a prompt regenerator rather than a prompt optimizer, and \ul{ we regard iterative-APE as an important baseline for evaluating the true effectiveness of LLM as prompt optimizers}.

    \noindent \textbf{APO} \cite{pryzant2023automatic}: A typical explicit reflection-based method. In each step, APO instructs the LLM optimizer to generate three reasons for the current errors and then generate new prompts based on the analyzed reasons.
    
    \noindent \textbf{PromptAgent} \cite{wang2023promptagent}: This method extends the APO paradigm by incorporating strategic reflection to introduce more expert-level prior into the task prompt.
    
    \noindent \textbf{APO-Sum} An APO extension introduced by this work. Inspired by \citet{sun2023autohint}, APO-Sum summarizes the reasons for current errors during reflection. Additionally, it leverages the advantage of chain-of-thought reasoning \cite{wei2022chain} by consolidating the two-step reflection and generation process into a single step.
    
    \noindent \textbf{OPRO} \cite{yang2023large}: A typical implicit reflection-based method. OPRO lets the LLM to generate new prompts that can enhance the test accuracy by implicitly considering a trajectory of historical prompts and their scores. Unlike APO-like methods, OPRO does not elicit the direction of prompt optimization.

We implemented all methods using GPT-4 as the LLM-based prompt generator $\mathcal{G}$ or optimizer $\mathcal{O}$. Our evaluations were conducted with two different target models $\mathcal{M}$, GPT-3.5-Turbo and Llama-2-70B-chat \cite{touvron2023llama}. The detailed implementations of each method are provided in Appendix \ref{sec:implementation}. It's worth noting that, by standardizing the settings across various methods (Table \ref{tab:unifiy_design}), \ul{we are specifically evaluating the prompt updating module of each method rather than their vanilla versions.}

\paragraph{Evaluation Tasks}
We evaluate the prompt optimization methods on four BigBench \cite{srivastava2023beyond} tasks: Object Counting, Navigate, Snarks, and Question Selection. Among these, Object Counting and Navigate are logical reasoning tasks, and Snarks and Question Selection are natural language understanding tasks. Detailed statistics are included in Appendix \ref{sec:data_split}.

\subsection{Results}

Figure \ref{fig:results_unifying} shows the results of different methods implemented under the unified settings. We make the following observations:
(1) Iterative-APE achieves comparable results with reflection-based methods in many tasks and generally outperforms OPRO. These results are inconsistent with our expectations, given that APE performs directionless resampling rather than targeted prompt optimization. 
(2) Another interesting observation is that the extended versions of APO, i.e., PromptAgent and APO-Sum, do not notably improve over APO when implemented with identical search strategies. This is also counterintuitive: as our assumption suggests that designing more sophisticated reflection processes should lead to more comprehensive analyses of current prompts and introduce greater prior knowledge about the task, thus yielding better results. 
These results, combined with observation (1), raise questions about the underlying mechanism of LLM prompt optimizers. \ul{How does the behavior of LLM prompt optimizers influence the performance of explicit reflection-based methods? Does the LLM optimizer struggle to provide sound reflection, or does it fail to appropriately refine prompts based on the generated feedbacks? }

We can also notice that OPRO performs less effectively than other methods across most tasks (within the search size). This result might be because: 1) Although OPRO guides the LLM optimizer to generate better prompts based on the historical prompt trajectory, the results indicate that the LLM optimizer may struggle to comprehend what constitutes ``better prompts". In contrast, explicit reflection-based methods elicit specific directions for prompt refinement, leading to more improvements. 2) Compared with APE, OPRO samples new prompts within a less restricted space. Consequently, the LLM optimizer may not always sample around the most optimal prompts. In fact, \citet{yang2023large} also shows that OPRO's performance can be influenced by the order of historical prompts, indicating that the LLM optimizer may not proactively grasp the principle behind ``good prompts".

\paragraph{Takeaways} from the observations in this section:
\begin{enumerate}[leftmargin=1em, itemindent=0em, itemsep=0.1em, topsep=0.6em]
    \item The LLM optimizer might not be a good implicit reflection-based prompt optimizer.
    \item The underlying mechanism of explicit reflection-based prompt optimization may differ from our expectations. 
\end{enumerate}
In the next sections, we respectively delve into the reflection (\S\ref{sec:reflection}) and prompt refinement (\S\ref{sec:refinement}) processes of explicit reflection-based prompt optimization to address the raised questions.

\section{Did the LLM-based Prompt Optimizers Perform Valid Reflection?}\label{sec:reflection}

The results in \S \ref{sec:main_evaluation} raised questions about the underlying mechanism of explicit reflection-based optimizers. In this section, we delve into the reflection process, trying to answer the question:  Can LLM-based prompt optimizers effectively reflect on the error examples and current prompts, thus generating sound feedbacks for prompt refinement?

\begin{figure}

\centering
\subfloat{\includegraphics[width=0.5\linewidth]{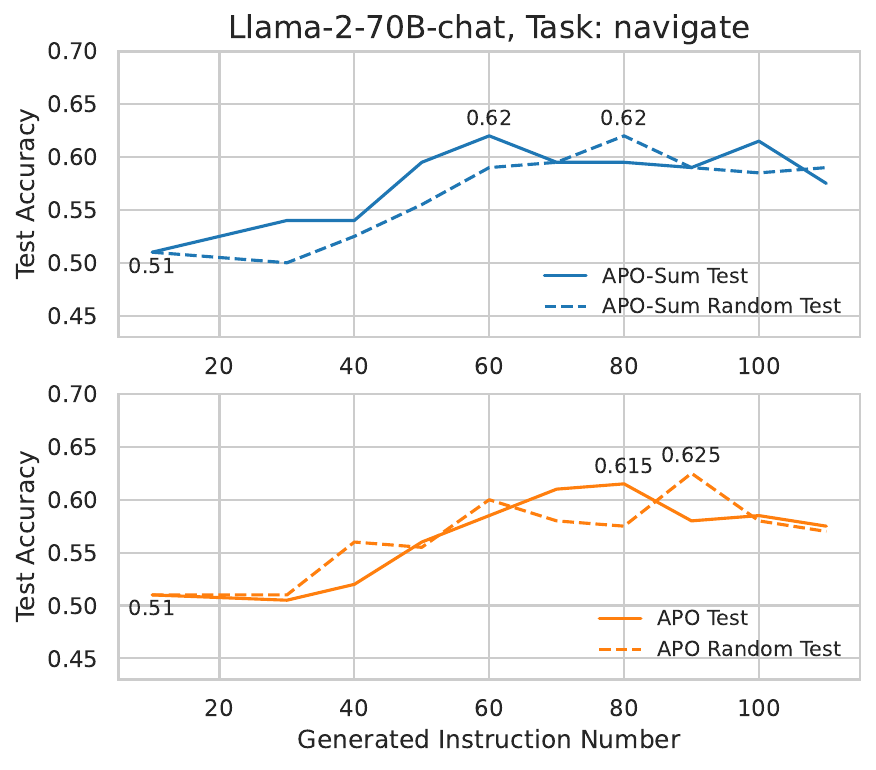}}
\subfloat{\includegraphics[width=0.5\linewidth]{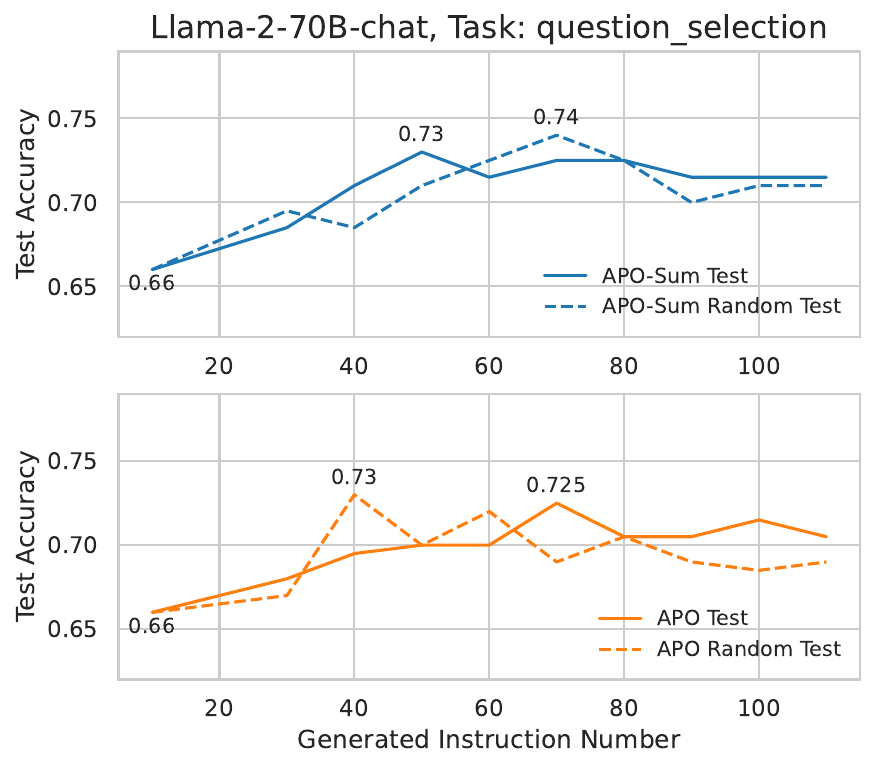}}\\
\subfloat{\includegraphics[width=0.5\linewidth]{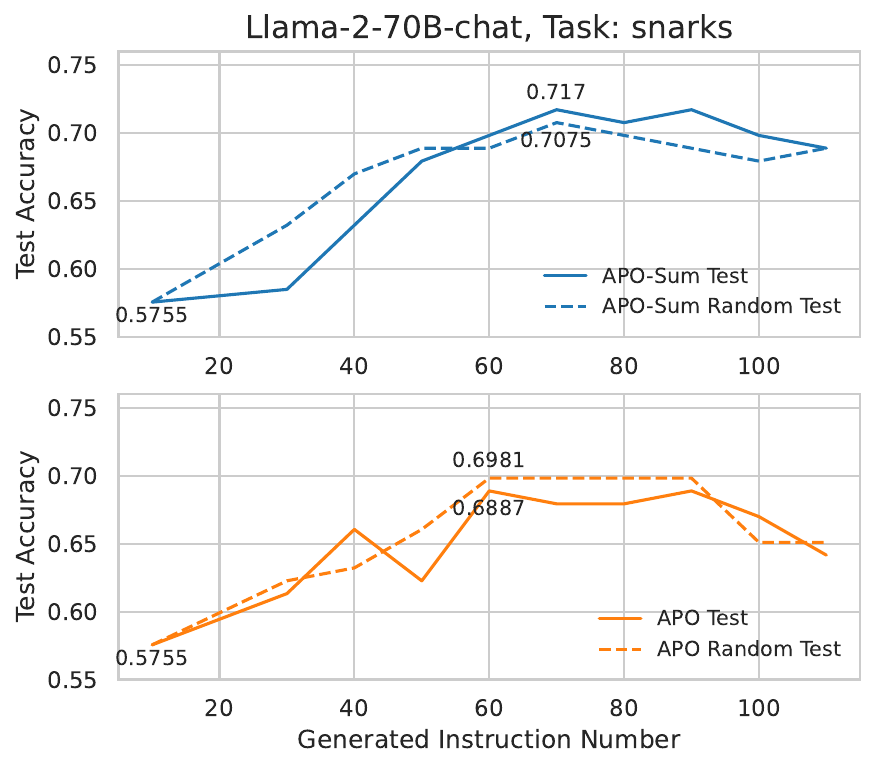}}
\subfloat{\includegraphics[width=0.5\linewidth]{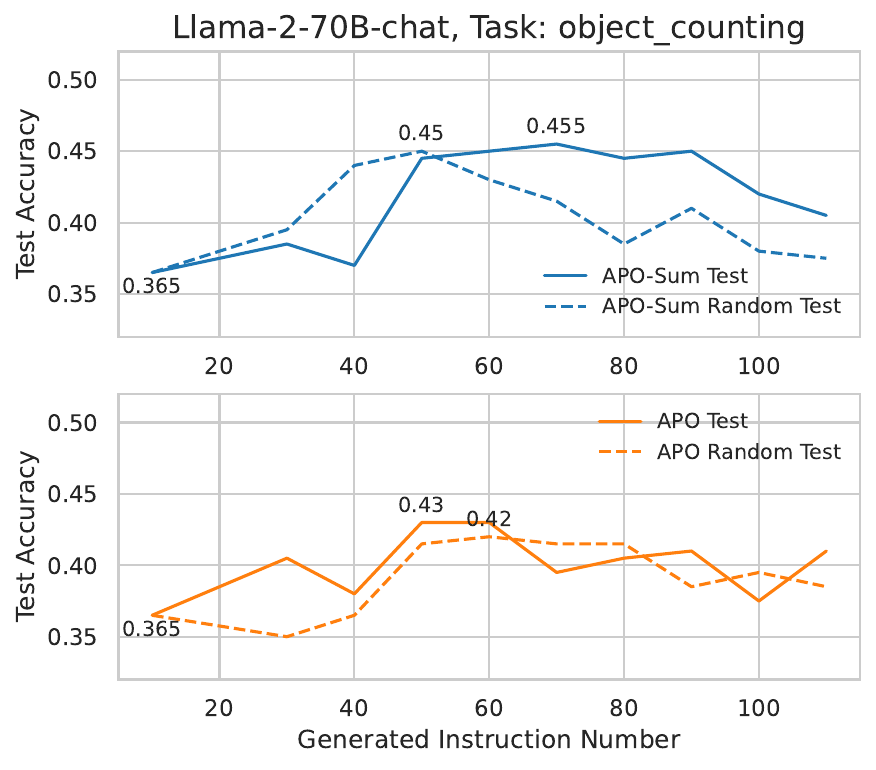}}
    \caption{ Comparison of Vanilla and Pseudo reflection settings on Llama-2-70B-chat.}
    \label{fig:random_error}
    \vspace{-0.2cm}
\end{figure}

\subsection{Experiment I: Reflecting on Pseudo Error}\label{sec:pseudo_error}
We first conduct an ablation study to examine the validity of the reflection process. We compare two settings for implementing the reflection process:

    \noindent \textbf{Vanilla}: The standard reflection setting where we sample a set of error examples $\mathcal{E}_{t-1}$ from the error distribution obtained 
    with the current prompt $p_{t-1}$. Here, ${E}_{t-1}=\{(x_i, y_i, a_i)\}_{i=1}^n$, each $(x_i, y_i, a_i)$ is an ``(input, gold answer, answer)" triple. Then, ${E}_{t-1}$ is included in the LLM-based reflection process by $c_{t-1} = \mathcal{O}(p_{t-1}, {E}_{t-1})$, where $c_{t-1}$ is the generated feedback.
    
    \noindent \textbf{Pseudo}: This setting replaces real error examples with pseudo error examples generated as follows: (1) We uniformly flip the LLM's predictions for all training examples, creating $\overline{a}_i$ for each example; (2) We uniformly sample a set of pseudo error examples $\overline{{E}}_{t-1}$ from the altered examples. These pseudo error examples $\overline{{E}}_{t-1}$ are then used for reflection generation: $\overline{c}_{t-1} = \mathcal{O}(p_{t-1}, \overline{{E}}_{t-1})$.

Except for the reflection process, all other aspects of the prompt optimization process are implemented identically for \textbf{Vanilla} and \textbf{Pseudo}. It is worth noting that when the target model $\mathcal{M}$ is Llama-2-70B-chat, each example's answer $a$ typically contains an extra explanation (see cases in Appendix \ref{sec:implementation}, Table \ref{tab:APO_case}). Therefore, it is expected to further support the LLM-based reflection.

\subsubsection{Results}

Figure \ref{fig:random_error} displays the results of APO and APO-Sum on Llama-2-70B-chat under different reflection settings. Surprisingly, we observe that the results, including the optimization trend, of Vanilla and Pseudo show comparability on tasks navigate, question\_selection, and snarks. On the object\_counting task, Pseudo exhibits slightly lower performance, yet the highest test scores are still close.
From these results, we can deduce that the LLM-based prompt optimizer may not perform valid reflections as we expect. \ul{Regardless of reflecting on the true error distribution or the pseudo error distribution, the LLM optimizer may generate similar feedback, resulting in comparable prompt optimization performance. It also suggests that the LLM-based prompt optimizers might struggle to identify the true issues with the current prompt based on the error examples.}

\begin{figure*}[ht]
    \centering
    \includegraphics[width=0.95\linewidth]{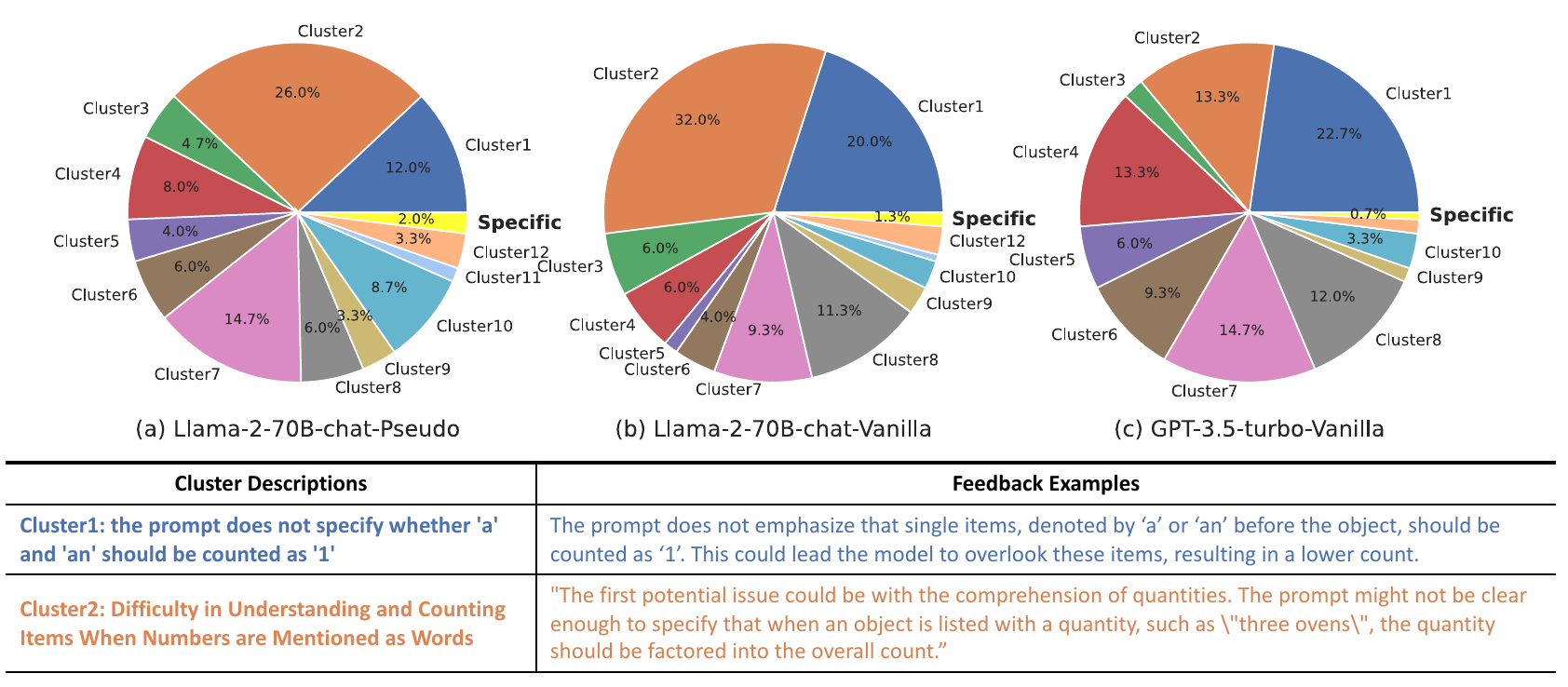}
    \caption{The feedback distribution of different methods, obtained by GPT-4-based feedback clustering. }
    \label{fig:feedback_distribution}
    \vspace{-0.2cm}
\end{figure*}

\subsection{Experiment II: Investigating the Feedback Distribution}

To validate the above assumptions, we further conducted experiments to analyze the distribution of generated feedbacks under different reflection settings. The questions are: (1) Did the LLM optimizer consistently generate similar feedbacks regardless of the error examples? (2) What types of feedbacks were most frequently generated, i.e., which issues about the prompts were most frequently discussed? Are these commonly raised issues valid?

\subsubsection{Feedback Clustering based on GPT-4}
To analyze the feedback distribution, an intuitive approach is to categorize feedbacks discussing similar issues into the same cluster. Inspired by recent research on LLM-as-a-judge \cite{zheng2023judging,bubeck2023sparks,dubois2023alpacafarm,zhou2023lima} and as a guide for text clustering \cite{zhang-etal-2023-clusterllm,an2023generalized,viswanathan2023large}, we developed a GPT-4-based clustering algorithm for analyzing feedback distribution. Specifically, at each step, GPT-4 is provided with a set of predefined clusters, each accompanied by a detailed cluster description, along with a batch of feedbacks. The model's task is to determine if each feedback belongs to any of the predefined clusters. If a feedback does not fit into any of the existing clusters, a new cluster should be created, complete with a fine-grained description. As the clusters are auto-created and may exhibit randomness, it's crucial to align the clusters from different feedback distributions for a fair comparison. To this end, we aggregate the feedbacks from various sources (to be compared) and input them together for the clustering process.
In Appendix \ref{sec:ap_clustering}, we present the clustering algorithm and provide details about the prompts used during the clustering process.

\subsubsection{Clustering Results}
Figure \ref{fig:feedback_distribution} shows the results of feedback clustering on the object\_counting task. We compare the feedback distributions obtained with three different settings, corresponding to \textbf{three different error distributions}: (a) the Pseudo reflection setting based on Llama-2-70B-chat (Llama-2-70B-chat-Pseudo); (b) the Vanilla reflection setting based on Llama-2-70B-chat (Llama-2-70B-chat-Vanilla); (c) the Vanilla reflection setting based on GPT-3.5-Turbo (GPT-3.5-turbo-Vanilla).
The pie charts depict the proportion of each cluster in the respective feedback distribution. The "Specific" category in each setting includes all specific types of feedbacks that did not appear in the other settings.

Our observations from the results are as follows:
(1) The feedback categories in different settings largely overlap. The major clusters have similar proportions, while the specific categories in each setting generally cover less than 2.0\% of the feedbacks. Among the distributions, the feedback distribution of Llama-2-70B-chat-Pseudo is similar to that of Llama-2-70B-chat-Vanilla, which \ul{validates our assumption that the LLM optimizer generates similar feedback regardless of the error distribution.}
(2) What types of issues are most frequently discussed? In the table below, we provide cluster descriptions and feedback examples for the two largest clusters. For instance, feedbacks in "Cluster1" discuss the issue that the model might not correctly count "a" or "an" as "1", constituting 12\%, 20\%, and 22.7\% of the feedback populations in the three settings, respectively. However, could this issue be a true reason for the errors that exist simultaneously in both Llama-2-70B-chat and GPT-3.5-Turbo? Furthermore, we present cases in Appendix \ref{sec:ap_clustering}, Table \ref{tab:repetitive_case}, showing that even when the prompt already included clear guidance on the ``"a" and "an" counted as 1'' issue, the LLM optimizer still raised the same concern (similar phenomena also occurred with other clusters). These results suggest the possibility that the LLM optimizer may struggle to identify the true reasons for errors. \ul{The LLM-based reflection process appears more like the LLM making educated guesses about the causes of errors based on its prior knowledge of the task, rather than genuinely reflecting and uncovering the true reasons.} 


\subsubsection{Instance-level Feedback Repetition}
We further examine the instance-level feedback repetition based on the clustering results. Specifically, we collect all the historical feedback of each single prompt during its evolution. Our main focus is to determine, at each step, the number of newly generated feedbacks that are repetitive with the prompt's historical feedbacks. To this end, we define $f_i$ as the $i^{th}$ generated feedback and $N_{f}$ as the total number of generated feedbacks for all prompts in the current step. We then calculate the Average Step Repetition Rate ($ASRR$) with $ASRR=\frac{\sum_i^{N_f}\mathcal{I}(f_i)}{N_{f}}$, where $\mathcal{I}(\cdot)$ is an indicator function indicating whether $f_i$ is repetitive with its corresponding historical feedbacks (i.e., whether $f_i$ belongs to the same cluster with at least one of its historical feedbacks).

\begin{figure}
    \centering
    \includegraphics[width=0.9\linewidth]{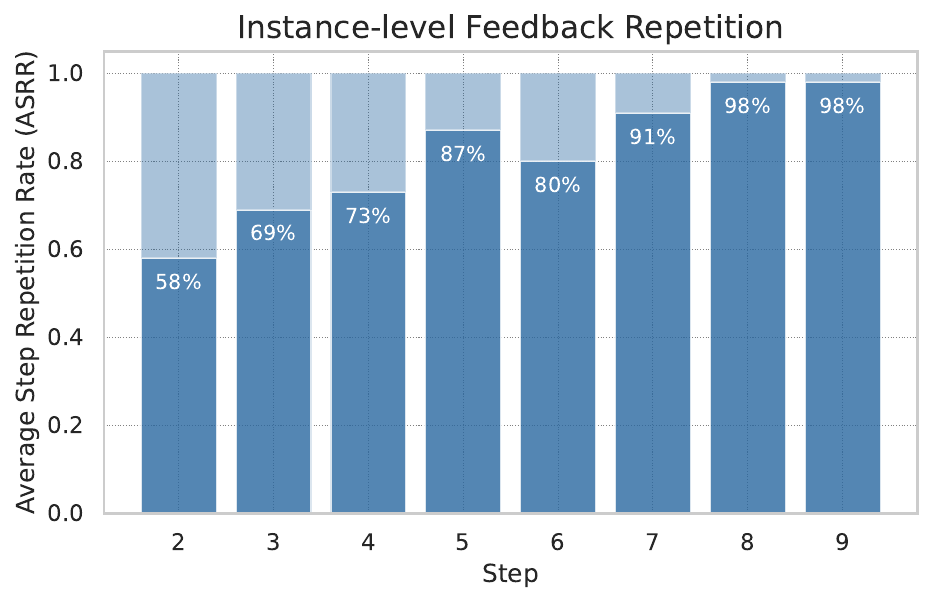}
    \caption{Instance-level feedback repetition results.}
    \label{fig:instance_repetition}
    \vspace{-0.3cm}
\end{figure}

In Figure \ref{fig:instance_repetition}, we present the results of $ASRR$ in each step. Starting from step 2, the $ASRR$ consistently remains above 50\% during the prompt optimization steps. This suggests that for each prompt, more than half of the issues raised during reflection have already been discussed, and are expected to have been resolved. These findings, on one hand, align with the assumption that the LLM optimizer may be speculating about the reasons for errors rather than uncovering the true reasons. On the other hand, \ul{since the true validity of the feedbacks remains uncertain, it is also possible that the LLM optimizer has indeed pointed out true reasons for the errors, yet it fails to appropriately refine the prompts based on the reflection.} 

\paragraph{Takeaways} from the findings in this section:
\begin{enumerate}[leftmargin=1em, itemindent=0em, itemsep=-0.2em, topsep=0.6em]
    \item The LLM optimizer generates similar feedbacks regardless of the error distributions.
    \item The LLM optimizer might be making educated guesses about the causes of errors based on its own prior knowledge rather than genuinely reflecting on the errors. 
    \item It is also possible that the LLM has indeed identified authentic problems in certain feedbacks, while it failed at the prompt refinement process.
\end{enumerate}

To figure out the actual situation, in the next section, we delve into assessing the validity of the prompt refinement process and its connection to the behavior changes of the target models.

\section{How is the Quality of the Refined Prompts and How They Affect the Target Models' Behavior?}\label{sec:refinement}

In Section \ref{sec:reflection}, we observed that the feedbacks generated by the LLM optimizer during reflection are largely repetitive and might not be sound. 
However, we also pointed out the possibility that the LLM optimizer may provide useful information in certain reflection steps (which does not conflict with the repetition), yet it failed to perform appropriate prompt refinement. It is important to know \ul{whether the unexpected results of reflection-based prompt optimization are due to the complete invalidity of the reflection process or the failure of prompt refinement.} To figure out this question, in this section, this section delves deeper into assessing the quality of refined prompts from various perspectives.

\begin{figure*}
    \centering
    \includegraphics[width=0.9\linewidth]{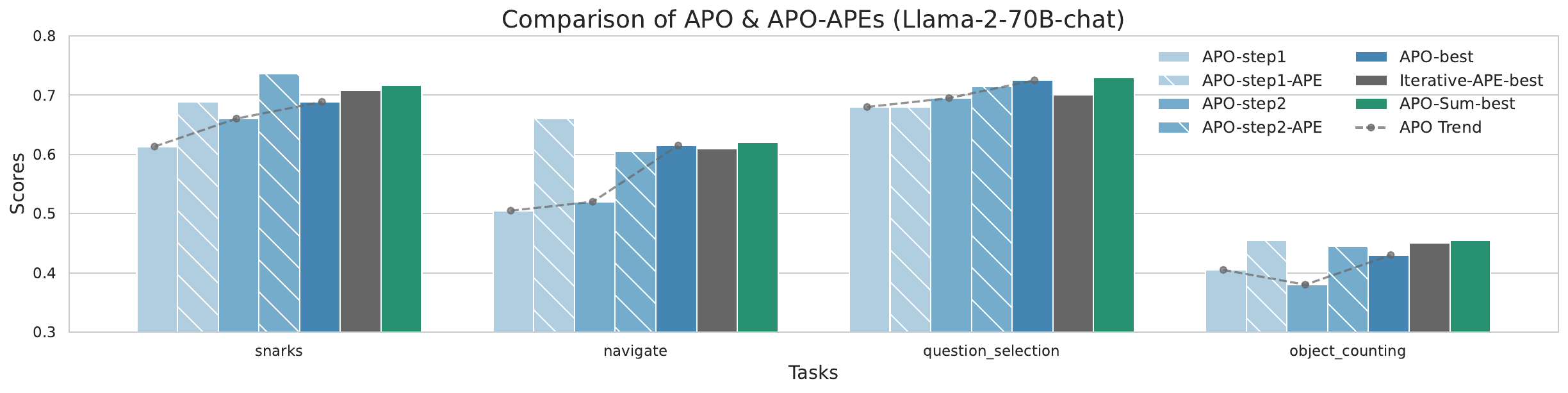} \\
    \includegraphics[width=0.9\linewidth]{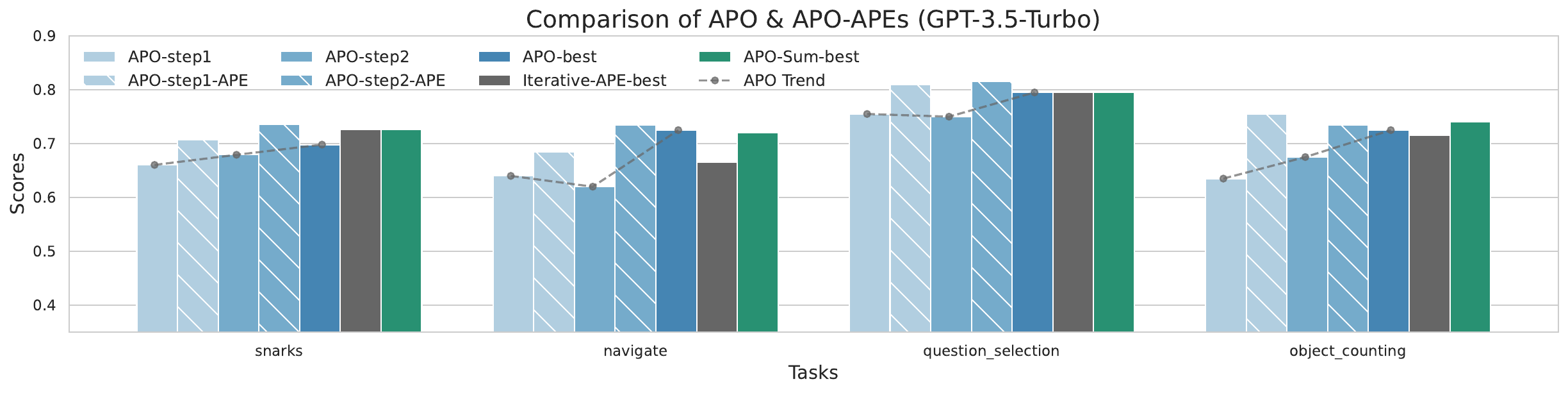}
    \caption{The results of performing Iterative-APE on APO-refined prompts. Performing Iterative-APE allows us to explore the most suitable prompts within the local semantic spaces, providing a fairer assessment of APO's semantic alterations to prompts.}
    \label{fig:stepAPE}
    \vspace{-0.3cm}                                   
\end{figure*}

\subsection{Experiment I: Evaluating the Refined Prompts from a Semantic View}\label{semantic}
The primary distinction between the resampling-based methods and the reflection-based methods lies in the alteration of the semantic meanings of prompts, with the latter modifying semantic meanings while the former does not. Consequently, evaluating the true quality of reflection and prompt refinement necessitates a fair evaluation of the altered semantic meanings—specifically, whether the alterations result in improvements compared to the original semantic content.

To achieve this goal, we applied iterative-APE to the APO-refined prompts in Step 1 and Step 2 (${P}_1$ and ${P}_2$ as defined in Section \ref{sec:search}), which means we collect the prompts selected in APO Step 1 (or Step 2), and apply iterative-APE to these prompts for following updating process.
By performing iterative-APE, we aim to explore the most suitable prompt within the local space of each altered prompt. This allows us to objectively assess the true validity of the semantic alterations made by APO, while minimizing the influence of other factors such as language, phrasing, and any randomness introduced during prompt sampling from the natural language space surrounding the desired semantic meanings.

\subsubsection{Results}
Figure \ref{fig:stepAPE} shows the best test scores of different methods. We denote the results of iterative-APE performed on APO Step 1 selected prompts $P_1$ as ``APO-Step1-APE'', and the results of iterative-APE performed on $P_2$ as ``APO-Step2-APE''. We also present the original test scores in APO Step 1 and APO Step 2, denoted as ``APO-Step1'' and ``APO-Step2''. ``APO-best'', ``APO-Sum-best'' and ``APE-best'' represent the best test scores of the corresponding methods. It's worth noting that ``APE-best'' also represents the best test scores of performing iterative-APE on the initial prompts $P_0$ (Step 0).
We also mark the trend of APO's original optimization process from Step 1 to Step 2, and ultimately to the best results.

Firstly, we can observe that the best performance among APO-Step1-APE and APO-Step2-APE consistently outperforms the performance of APE-best in most tasks. \ul{These results suggest that APO is able to make valid semantic alterations during certain reflection processes in Step 1 and Step 2}, which means the LLM prompt optimizer can indeed introduce useful information into the prompts through reflection in certain steps, thus expanding the upper bound of the prompting performance.

Secondly, the best performance among APO-Step1-APE and APO-Step2-APE also surpasses APO-best in most tasks. Besides, APO-Step1-APE and APO-Step2-APE notably improve upon APO-Step1 and APO-Step2 in most tasks, respectively. These results show that while APO is able to improve the semantic meanings of the prompts, \ul{most of the time, the LLM-based prompt optimizer fails to generate an appropriate prompt around the semantic space with a single refinement.} This observation may also provide an explanation for the phenomenon in Figure \ref{fig:results_unifying}, where different reflection-based methods achieve similar results regardless of their designs. This suggests that a fairer comparison of these methods by evaluating the refined semantic spaces might yield different outcomes. More importantly, \ul{it highlights the potential of performing fine-grained search when prompts already contain sufficient semantic information}. In fact, it's evident that the semantic refinement achieved through a single optimization step may be adequate for most tasks, as seen in the results of APO-Step1-APE.

Another noteworthy observation is that the performance of APO-Step2-APE is not consistently better than APO-Step1-APE. This suggests that \ul{the semantic alterations made by reflection-based prompt optimizer is not always valid}. These observations correspond with the repetitive feedbacks and the similar feedbacks generated across different error distributions in \S \ref{sec:reflection}. In fact, we want to clarify that the overall invalidity of repetitive reflections and the semantical validity of certain reflections are not contradictory. \ul{The invalid reflection problem and the inappropriate prompt refinement problems may co-exist in LLM-based prompt optimization, simultaneously leading to the inefficiency of existing methods.}

\subsection{Experiment II: Investigating the Instruction Following Behaviors of the Target Models}\label{sec:following_experiment}

In the last section, we deduced that the LLM prompt optimizers struggle to generate appropriate prompts even based on valid semantic content. Such an issue is also closely related to the behavior of the target models. Prior research has widely explored the LLM's prompt sensitivity \cite{webson-pavlick-2022-prompt,min2022rethinking,lu-etal-2022-fantastically,zhao2021calibrate}, its challenges in handling lengthy context \cite{liu2023lost,li2023how}, and its limited capability to strictly follow instructions \cite{zhou2023instruction,zeng2023evaluating}. All these factors may result in a gap between LLM-based prompt generation and the actual performance of the prompts on the target models.

In this section, we delve deeper into the problem of LLM-based prompt refinement from the perspective of target model behavior. Our key inquiries are as follows: (1) When the LLM prompt optimizer introduces new prior information or guidance into the prompts (as illustrated in cases presented in Table \ref{tab:APO_case}, \ref{tab:APO_sum_case}), is this new guidance or information comprehended and effectively utilized by the target models? (2) Does the introduction of new information affect the models' attention to existing information?

To this end, we selected two ``verifiable instructions'' following \citet{zhou2023instruction}:
\begin{itemize}[leftmargin=1em, itemindent=0em, itemsep=-0.3em, topsep=0.4em]
    \item \textit{Include the keyword "Alright" in your response.}
    \item \textit{First repeat the input without change, then give your answer.}
\end{itemize}
These two instructions allow us to objectively assess whether the target models have followed the instructions. For example, the following to the first instruction can be evaluated by extracting the word ``Alright'' from the response. 

\begin{figure}
    \centering
    \includegraphics[width=1.0\linewidth]{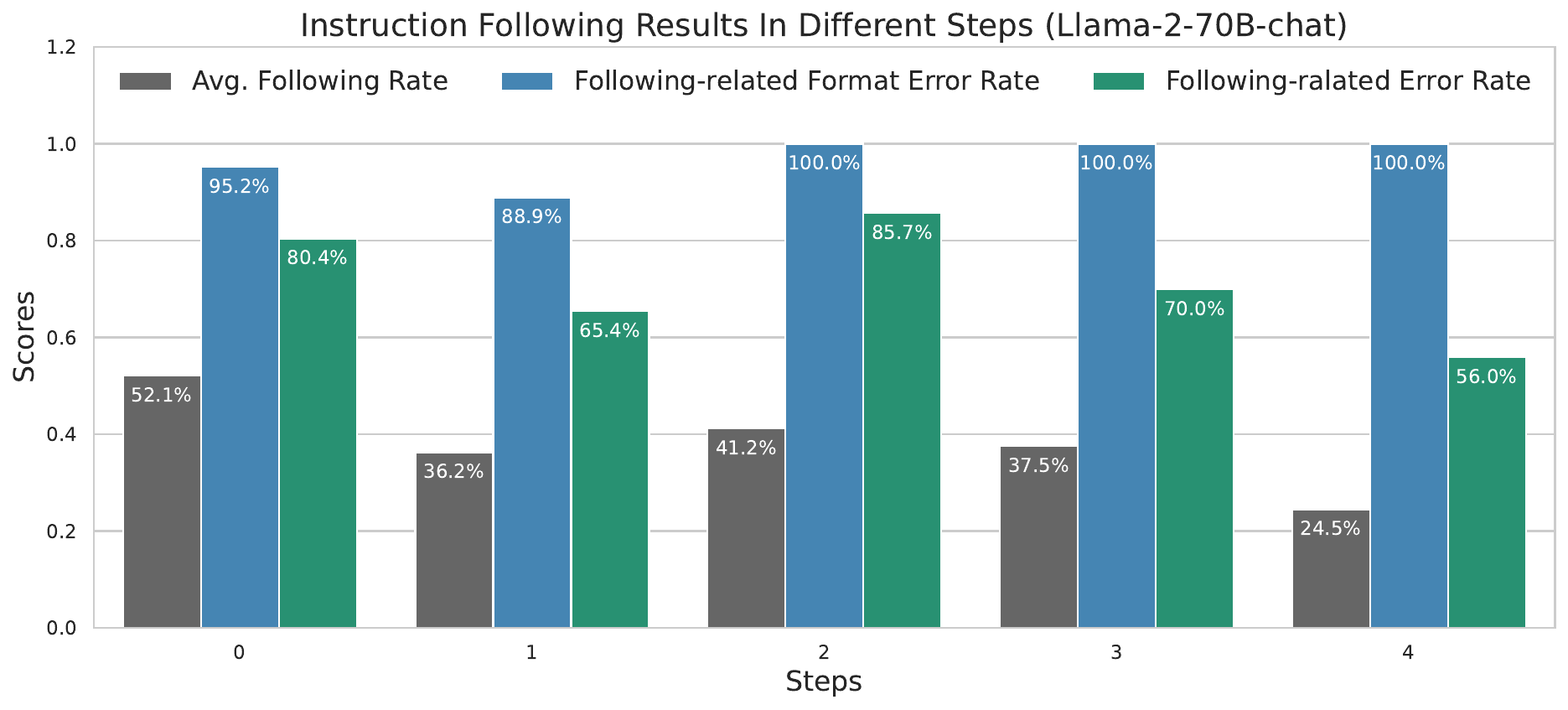} \\
    \includegraphics[width=1.0\linewidth]{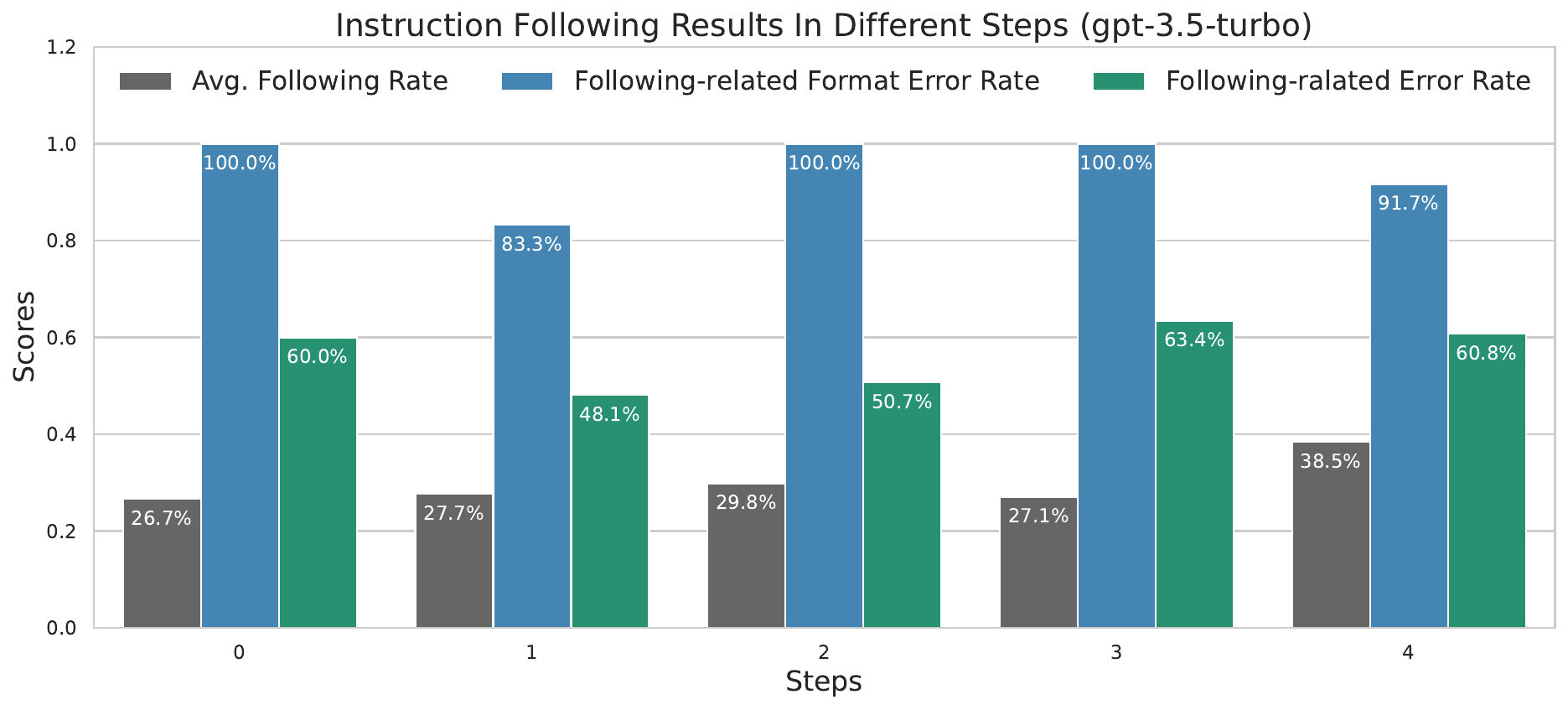}
    \caption{The instruction following results.}
    \label{fig:follow_result}
    \vspace{-0.2cm}                                   
\end{figure}

Next, we instructed the LLM optimizer to ``insert the instruction at an appropriate position'' in the prompts from each step of APO. Then, for each prompt, we evaluate the model's behavior with the original and modified prompts. We introduce the following metrics ( formal definitions of each metric are presented in Appendix \ref{sec:ap_following}):
\begin{itemize}[leftmargin=1em, itemindent=0em, itemsep=0em, topsep=0.6em]
    \item Average Following Rate ($AFR$): This metric describes the average ratio of successful guidance-following (successfully following the inserted guidance) when enumerating all the inputs in the training set. 
    \item Following-related Format Error Rate ($FFER$): This metric measures the ratio of successful guidance-following prompts among the prompts that exhibit poorer following scores regarding the format requirement. It describes the impact of the target models adhering to new guidance on following existing guidance.
    \item Following-related Error Rate ($FER$): Similar to $FFER$, this metric measures the ratio of successful guidance-following prompts among the prompts that yield lower scores on the task. It describes the impact of incorporating additional task-unrelated guidance on task performance.
\end{itemize}


\subsubsection{Results}
In figure \ref{fig:follow_result}, we show the instruction-following results on APO Step $0$ to Step $4$. Each reported result is the average result of inserting ``repeat'' and ``keyword'' instructions on navigate and object\_counting tasks (average results of 4 settings). 

The results show that the Average Following Rate of Llama-2-70B-chat is consistently below 50\% in most steps, and there is a declining trend as the steps progress. This suggests that the model encounters difficulties in following new guidance, especially as prompts become longer or already contain a certain amount of guidance. In contrast, GPT-3.5-Turbo exhibits a different behavior from Llama-2-70B-chat. It maintains a generally consistent $AFR$ across different steps, but the $AFR$ consistently remains lower ($AFR<30\%$) in most steps. This indicates GPT-3.5-Turbo's weaker ability to follow instructions in certain cases, yet it appears to be more robust to prompt length.

Both models show high Following-related Error Rates ($FFER$), indicating that the introduction of new guidance significantly increases format errors of the target models. This phenomenon, combined with the observations before, suggests that \ul{when new guidance is introduced through reflection-based prompt refinement, the models may struggle to follow it or may shift their attention away from existing information while following new guidance. Both behaviors can lead to the ineffectiveness of the new prompts, rendering them "inappropriate''} (as discussed in Section \ref{semantic}). Such a gap between prompt generation and the target models' following behavior may potentially be mitigated by searching for improved wording or language within the semantic space that the target models can more easily follow (as indicated by the results in Figure \ref{fig:stepAPE}).

\paragraph{Takeaways} from the findings in this section:
\begin{enumerate}[leftmargin=1em, itemindent=0em, itemsep=0em, topsep=0.6em]
    \item The reflection-based prompt optimizers are able to make valid \textit{semantic alterations} on prompts in certain reflection steps.
    \item Nevertheless, the LLM optimizers struggle to generate appropriate prompts in the semantic spaces with just a single refinement.
    \item The invalid reflection and the inappropriate prompt refinement problems may co-exist in LLM-based prompt optimization, simultaneously leading to the inefficiency of existing methods.
    \item The challenge faced by LLM optimizers in generating suitable prompts can be partly attributed to the gap between prompt generation and the uncontrollable instruction-following behavior of the target models.
\end{enumerate}

\section{Large Language Models are Good ``Behavior Optimizers"}\label{sec:behavior}

In earlier sections, we discussed the problems of LLM as prompt optimizers during both the reflection and the prompt refinement process. These problems primarily arise from the gap between the LLM prompt optimizer and the target models. The LLM optimizer struggles to understand the failures of the target models, while the target models face difficulties in appropriately following the generated instructions. Therefore, we advocate for the exploration of new paradigms for LLM-based Prompt Optimization that can substantially alleviate these issues. 

\subsection{Automatic Behavior Optimization}
In this work, we take a preliminary step by introducing a new paradigm for LLM-based prompt optimization named ``Automatic Behavior Optimization'' (ABO). While existing LLM-based Prompt Optimization methods focus on optimizing prompts to \ul{affect the models' behavior}, Automatic Behavior Optimization aims to \ul{directly optimize the models' behavior}. 
This is achieved through the following steps:
\begin{enumerate}[leftmargin=1em, itemindent=0em, itemsep=0em, topsep=0.6em]
    \item At each step, the LLM optimizer is instructed to generate step-by-step prompts. 
    \item Next, we utilize the LLM optimizer to write an ``Instruction-following Demonstration'' for each prompt, i.e., an example illustrating how to strictly follow every detail of the given prompt. This practice aims to enhance the controllability of the prompt optimization process by ensuring any improvement made will be strictly followed by the target models.
    \item During the reflection and prompt refinement process, given error examples, the LLM optimizer is required to identify the failure step of the target model and refine the prompt by further breaking down the solution at the problematic step. This aims to avoid invalid feedback by utilizing the LLM optimizer to perform more objective tasks during reflection. 
    \item For each refined prompt, the instruction-following demonstration is also updated to illustrate how to strictly follow the refined steps.
\end{enumerate}

\begin{figure*}
    \centering
    \includegraphics[width=0.9\linewidth]{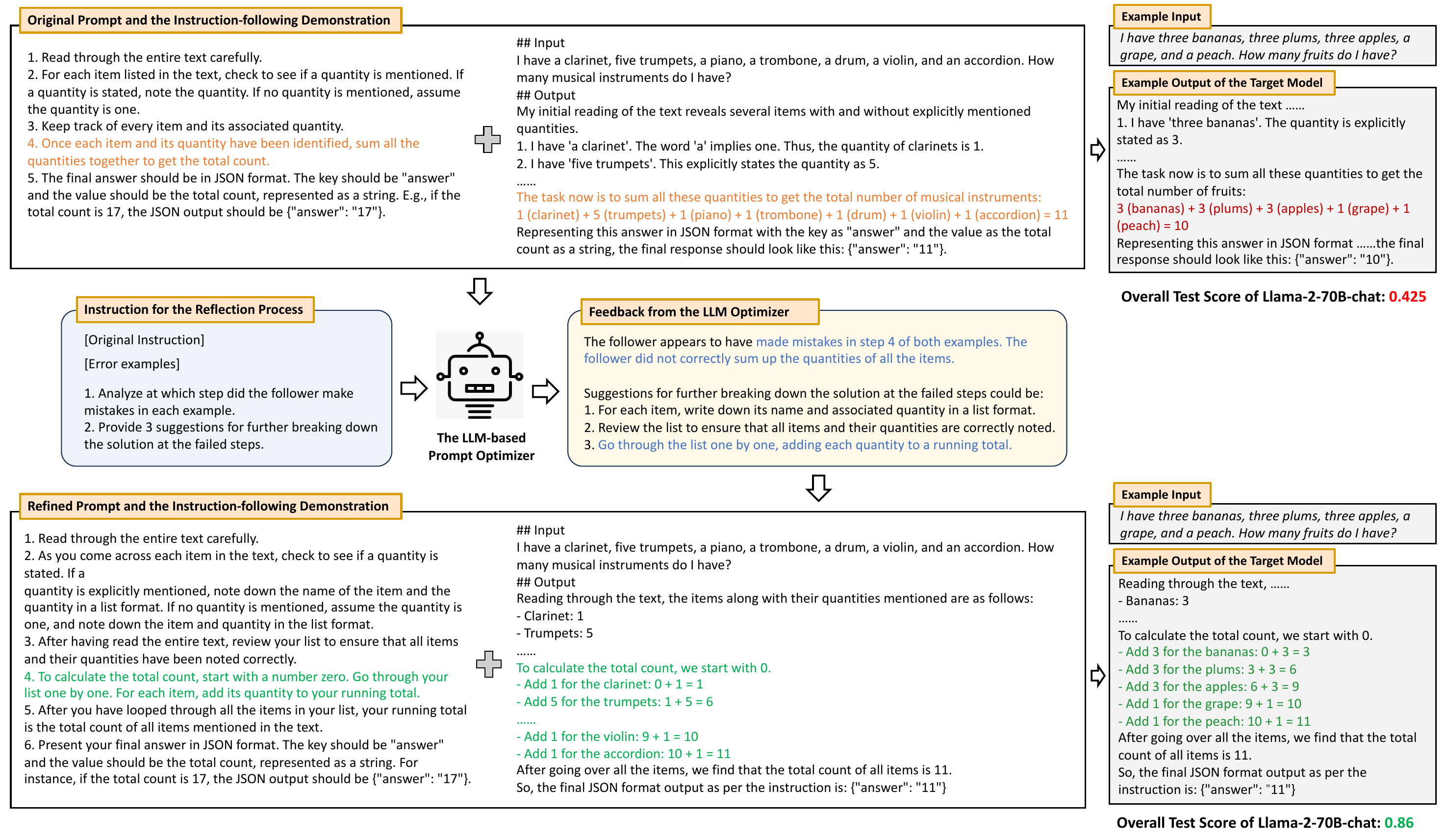} 
    \caption{Illustration and case study of the Automatic Behavior Optimization process. Example is taken from the Automatic Behavior Optimization process on object\_counting with Llama-2-70B-chat.}
    \label{fig:behavior_method}
    \vspace{-0.2cm}                                   
\end{figure*}

Figure \ref{fig:behavior_method} illustrates the process of the Automatic Behavior Optimization process with a case selected from the ABO process on object\_counting with Llama-2-70B-chat. We find that ABO is extremely effective when target models exhibited limited capability to fulfill the task. By continuously refining the target model's behavior at its weak points, ABO is able to find the most suitable behavior that best fits the target model's capability.

\subsection{Experimental Settings}
The implementation of ABO is also under the unified settings, except that we instructed the LLM optimizer to write an instruction-following demonstration for each initial prompt, which constructs the prompts in optimization Step 0. We compare ABO with the following baselines: (1) Although we add the instruction-following demonstrations mainly for controlling the target models' behavior, this practice is related to one-shot CoT prompting. Therefore, we include the zero-shot CoT \cite{kojima2022large} and the few-shot CoT prompts used in BBH \cite{suzgun-etal-2023-challenging} (3-shot CoT prompts) as two CoT-based baselines, denoted as Zero-shot-CoT and Few-shot-CoT, respectively. (2) We also include the best results of APO we have obtained in the previous sections (denoted as APO-All-best), which mostly come from the results of APO-APEs methods (Figure \ref{fig:stepAPE}). Note that these results may not be the best results of APO and its variations, since we have proved that by performing fine-grained search we can always obtain better results. We include this baseline only to make a coarse comparison of different paradigms.

\paragraph{Ablation Study} We also conducted an ablation experiment on ABO: Based on the best prompts obtained by ABO, we removed the instruction-following demonstration in the prompts. Instead, we use a ``Strictly follow every detail of the instruction.'' system prompt, and add a ``STRICTLY follow every detail of the following instruction.'' instruction before the input. We denote this method as ABO-Ablation, which aims to check, without the demonstration, to which degree can the target models adhere to the prompts and benefit from the best behaviors described in the prompts.

\vspace{-0.3cm}
\begin{table}[ht]
\setlength{\tabcolsep}{5pt}
\centering
\caption{Results of Automatic Behavior Optimization.}
\vspace{0.3cm}
\scalebox{0.73}{
\small
\begin{tabular}{lcccccccc}
\toprule

\multirow{2}*{\bfseries Methods} & \multicolumn{4}{c}{\textbf{Target Model: Llama-2-70B-chat}} \\ 
\cmidrule(r){2-5} 
& \makecell[c]{object\_counting} & \makecell[c]{navigate} &
\makecell[c]{snarks} & \makecell[c]{question\_selection} \\

\midrule
 Zero-shot-CoT &
0.425  & 0.645  & 0.651 & 0.695\\
 Few-shot-CoT &
 \ul{0.505}& \ul{0.720}& 0.547& 0.715 \\
  \cmidrule(r){1-5} 
 APO-All-best &
 0.455& 0.660& \ul{0.736}& \ul{0.725} \\
 \cmidrule(r){1-5} 
 ABO (Step 0) &
0.485& 0.610& 0.594& 0.705 \\
 ABO (Step 1) &
0.860& 0.845&\bfseries 0.811& 0.785 \\
 ABO (Step 2) &
\bfseries 0.885& \bfseries 0.890& 0.793& \bfseries 0.810 \\
  \cmidrule(r){1-5} 
 ABO-Ablation  &
0.385& 0.625& 0.670 & 0.680 \\

\bottomrule
\toprule
\multirow{2}*{\bfseries Methods} & \multicolumn{4}{c}{\textbf{Target Model: GPT-3.5-Turbo}} \\ 
\cmidrule(r){2-5} 
& \makecell[c]{object\_counting} & \makecell[c]{navigate} &
\makecell[c]{snarks} & \makecell[c]{question\_selection} \\

\midrule
 Zero-shot-CoT &
0.695  & 0.580  & 0.708 & 0.795\\
 Few-shot-CoT &
 \bfseries 0.985& \ul{0.925}& 0.642& \ul{0.860} \\
  \cmidrule(r){1-5} 
 APO-All-best &
 0.755& 0.735& \ul{0.736}& 0.815 \\
 \cmidrule(r){1-5} 
 ABO (Step 0) &
0.895& 0.810& 0.697& 0.740 \\
 ABO (Step 1) &
0.935& 0.965&\bfseries 0.802& 0.760 \\
 ABO (Step 2) &
 \ul{0.975}& \bfseries 0.985& 0.764& \bfseries 0.905 \\
  \cmidrule(r){1-5} 
  ABO-Ablation &
0.430 & 0.640 & 0.547 & 0.700 \\


\bottomrule

\end{tabular}
}
\label{tab:behavior_result}
\vspace{-0.2cm}
\end{table}

\subsection{Results}

Table \ref{tab:behavior_result} shows the results of ABO and the baseline methods. We can observe that:
(1) Compared to the results of ABO (Step 0), ABO can effectively improve the results in Step 1 and Step 2. As the prompts in ABO (Step 0) also include instruction-following demonstrations, these results verify that the superior performance of ABO comes from the refinement of target models' behavior, rather than the existence of demonstrations.
(2) The results of ABO generally outperform ABO-All-best, Zero-shot-CoT and Few-shot-CoT in most tasks, and show significant improvement on the object\_counting and navigate tasks on Llama-2-70B-chat. These results are rather interesting and can be demonstrated with the case in Figure \ref{fig:behavior_method}. The case shows that Llama-2-70B-chat struggles to correctly sum up number lists, probably due to a lack of arithmetic ability.
However, ABO broke down the summation step into ``calculate one by one'' and ensured strict instruction-following of the target model, leading to significant improvement. Additionally, \citet{suzgun-etal-2023-challenging} considered snarks as a task that CoT-prompting failed on LLMs (which can be illustrated by the Few-shot-CoT results). However, ABO is able to find suitable behavior for each model on snarks, leading to superior results than human-written CoT. \ul{These results indicate that ABO-based behavior improvement is rather effective when the target models' capability used to be inadequate for the task.}

The results of ABO-Ablation are also interesting. Although ABO has found the optimal behavior of the target models, without the instruction-following demonstrations to control the target models' behavior, the target models still fail to benefit from the prompts, even with the ``Strictly following'' instructions. These results correspond with the observations in \S \ref{sec:following_experiment} that \ul{the gap between prompts and the target models' behavior is a significant problem in LLM-based prompt optimization, and ensuring strict following behavior of the target models can largely enhance the efficiency of prompt optimization.}

\section{Conclusion}
In this work, we conducted a comprehensive study to uncover the underlying mechanism of LLM-based Automatic Prompt Optimization. We first isolate the effect of various LLM-based prompt optimizers with a unified setting, showing that the behaviors of LLM optimizers differ from our expectations. Next, we respectively delve into the reflection and prompt refinement process of reflection-based prompt optimization, demonstrating the ineffectiveness of LLM optimizers during both processes. These observations advocate for the exploration of new paradigms in LLM-based prompt optimization, where we take a preliminary step by introducing a new Automatic Behavior Optimization paradigm. We hope our study can inspire the development of more new paradigms and further work in the field of LLM-based automatic prompt optimization.

\newpage

\nocite{langley00}

\bibliography{anthology,custom}
\bibliographystyle{icml2024}

\newpage
\appendix
\onecolumn

\section{Implementation Details}\label{sec:implementation}

We implemented all LLM-based Prompt Optimization methods based on the APE repository \footnote{\url{https://github.com/keirp/automatic_prompt_engineer}}. During the initialization stage, we utilized $4$ initial examples for LLM-based prompt initialization and sampled a total number of $2 \times 5$ initial prompts with $2$ queries (each sample $5$ prompts). For all Reflection-based methods (APO, PromptAgent, APO-Sum, and OPRO), we sampled $4$ error examples for the reflection process of each prompt. For other implementation details for OPRO, we adhered to the default settings in the paper, which used a history record of $20$ prompts arranged in ascending order for prompt updating. We present all the used prompts and prompting cases for each method in Table \ref{tab:APO_meta} to Table \ref{tab:OPRO_case}.

We used GPT-4\footnote{\url{http://openai.com/api}} for prompt initialization and prompt updating for all methods, with default hyperparameters except for a temperature of $0.9$. We utilized GPT-3.5-Turbo\footnote{\url{http://openai.com/api}} and Llama-2-70B-chat\footnote{\url{https://huggingface.co/meta-llama/Llama-2-70b-chat-hf}} as target models, both setting the temperature to $0$ for inference, following previous work \cite{yang2023large}. To accelerate the inference of Llama-2-70B-chat, we implemented it with the vLLM inference library \cite{kwon2023efficient}\footnote{\url{https://github.com/vllm-project/vllm}}.

\paragraph{Dateset Split}\label{sec:data_split}
The detailed statistics of the used datasets are shown in Table \ref{tab:dataset_detail}. We filtered 15 instances in snarks involving the hate, sexual, violence, and self-harm categories that were detected by the OpenAI content filter \footnote{\url{https://learn.microsoft.com/en-us/azure/ai-services/openai/concepts/content-filter}}.

\begin{table}[h]
\centering
\renewcommand\arraystretch{1.6}
\scalebox{0.7}{
\begin{tabular}{|l|c|c|c|c|c|c|c|c|c|c|c|}
\hline
\bfseries Dataset Name &\bfseries Initialization &\bfseries Train \& Dev &\bfseries\ \ \ \ \  Test \ \ \ \ \  \\
\hline
Object Counting & 10 & 50 & 200 \\
Navigate & 10 & 50 & 200 \\
Snarks & 10 & 50 & 106 \\
Question Selection & 10 & 50 & 200 \\
\hline

\end{tabular}}
\caption{Statistics of the used datasets. }
\label{tab:dataset_detail}
\vspace{-0.3cm}
\end{table}

\section{Additional Details for the Feedback-clustering Experiment}\label{sec:ap_clustering}
In Algorithm \ref{alg:cluster}, we show the algorithm used for GPT-4-based feedback clustering. In Table \ref{tab:cluster_prompt}, we show the prompt used for querying GPT-4 for clustering. In Table \ref{tab:cluster_1} to Table \ref{tab:cluster_3}, we show the clustering results on object\_counting, including the description of each cluster and feedback examples belonging to the cluster. 

\paragraph{Cases for Instance-level Repetition} In Table \ref{tab:repetitive_case}, we show the cases of instance-level feedback repetition. When the current prompts already include clear guidance on the concerned issues, the LLM optimizer still generated feedbacks that discussed the same issues.

\section{Additional Details for the Instruction-following Experiment}\label{sec:ap_following}
\subsection{Metric Definition}
In this section, we define the metrics used in Section \ref{sec:following_experiment}. Formally, we denote the number of prompts in Step $t$ as $N_{t}$. Each prompt is evaluated on the whole training set $\mathcal{D}_{tr}$. Then, we define the following metrics:
\begin{enumerate}[leftmargin=1em, itemindent=0em]
    \item Average Following Rate $AFR$: $AFR=\frac{\sum_i^{N_t} \sum_{x\sim \mathcal{D}_{tr}} \mathbb{I}_1(\mathcal{M}, x, {p}_t^{i,insert})}{N_t\cdot |\mathcal{D}_{train}|}$, where ${p}_t^{i,insert}$ is the altered prompt, $\mathcal{M}$ is the target model and $\mathbb{I}_1(\cdot)$ is an indicator function indicating whether $\mathcal{M}$ correctly follow the inserted instruction with input $x$.
    \item Following-related Format Error Rate ($FFER$): This metric measures to what degree the following to the extra instruction affects the following to the format requirement. We denote $\mathbb{I}_2(\mathcal{M},{p}_t^{i,insert})$ as whether $\mathcal{M}$ successfully followed the insert instruction in ${p}_t^{i,insert}$ on at least one example.
    Then, we calculate $FFER$ by: $FFER=\frac{\sum_i^{N_t} \mathbb{I}_2(\mathcal{M},{p}_t^{i,insert}) \cdot  \mathbb{I}_3(\mathcal{M}, \mathcal{D}_{tr}, p_t^i, {p}_t^{i,insert})}{\sum_i^{N_t} \mathbb{I}_3(\mathcal{M}, \mathcal{D}_{tr}, p_t^i, {p}_t^{i,insert})}$, where $\mathbb{I}_3(\cdot)$ is an indicator function indicating whether the number of format errors increases on $\mathcal{D}_{tr}$ when $p_t^i$ is changed to ${p}_t^{i,insert}$.

    \item  Following-related Error Rate ($FER$): This metric measures to what degree the following to the extra instruction affects the score. Similar to $FFER$, we define $FER$ by $FFER=\frac{\sum_i^{N_t} \mathbb{I}_2(\mathcal{M},{p}_t^{i,insert}) \cdot  \mathbb{I}_4(\mathcal{M}, \mathcal{D}_{tr}, p_t^i, {p}_t^{i,insert})}{\sum_i^{N_t} \mathbb{I}_4(\mathcal{M}, \mathcal{D}_{tr}, p_t^i, {p}_t^{i,insert})}$. Here, $\mathbb{I}_3(\cdot)$ is another indicator function indicating whether the score on $\mathcal{D}_{tr}$ degrades when $p_t^i$ is changed to ${p}_t^{i,insert}$.
\end{enumerate}

\subsection{Case Study}
In Table \ref{tab:following_case_llama_repeat} to Table \ref{tab:following_case_35_keyword}, we show cases of the instruction-following experiments. These cases illustrate how we insert the extra guidance and how the target models failed to be restricted to the answer format or answer correctly when following the new guidance.

\section{Additional Details for the Automatic Behavior Optimization Experiment}\label{sec:ap_ABO}
In Table \ref{tab:ABO_best_llama1} to Table \ref{tab:ABO_best_35_4}, we present the best-optimized prompts obtained with Automatic Behavior Optimization in $3$ steps.

\begin{table*}[h]
\centering
\begin{tabular}{p{5cm} p{8cm}}
\hline
Meta Format Name & Meta Format
\\
\hline
\\
reflection\_format&
\makecell[l]{
I'm trying to write a task prompt.\\\\
My current prompt is: \\
"
\texttt{[} ins \texttt{]}
"\\\\
But this prompt gets the following examples wrong: \\
\texttt{[} error\_examples \texttt{]}\\\\
give 3 reasons why the prompt could have gotten \\
these examples wrong. Wrap each reason with\\
$<$START$>$ and $<$END$>$.
}
\\
\hline
\\
error\_example&
\makecell[l]{
\#\#\# Sample \texttt{[}index\texttt{]}\\
\#\# Input\\
\texttt{[} input \texttt{]}\\
\#\# Correct Answer\\
\texttt{[} answer \texttt{]} \\
\#\# Output\\
\texttt{[}llm\_output \texttt{]}
}\\
\\
\hline 

\\
prompt\_refinement&
\makecell[l]{
I'm trying to write a task prompt.\\\\
My current prompt is: \\
"\\
\texttt{[} ins \texttt{]}\\
"  \\\\
But it gets the following examples wrong:\\
\texttt{[}error\_examples \texttt{]}\\\\
Based on these examples the problem with this \\
prompt is that :\\
\texttt{[} feedbacks \texttt{]}\\\\
Based on the above information, I wrote 2 different \\
improved prompts. Each prompt is wrapped with \\
$<$START$>$  and $<$END$>$. \\\\
The 2 new prompts are:
}\\
\hline
\end{tabular}
\caption{The meta prompt formats used for APO.}
\label{tab:APO_meta}
\end{table*}

\begin{table*}[h]
\centering
\scalebox{0.8}{
\begin{tabular}{p{3cm} p{13cm}}
\hlineB{3}

\multicolumn{2}{c}{Case of APO}
\\
\hline
\\
Original Prompt&
\tabincell{ p{13cm}}{
You are given a sequence of navigation instructions. Please interpret these instructions in a two-dimensional context (considering left, right, forward, and backward movements). Based on these instructions, provide a concise explanation of the path taken. Then, determine if the journey ends at the starting point. If yes, respond with "A"; if no, respond with "B". Your response should be in the following format: \{"answer": "A"\} or \{"answer": "B"\}.
}\\
\\
\hline\\
Error Examples&
\tabincell{ p{13cm}}{
\#\#\# Sample 1\\
\#\# Input\\
Turn around. Turn right. Take 5 steps. Take 2 steps.\\
\#\# Correct Answer\\
\{"answer": "B"\}\\
\#\# Output\\
\{"answer": "A"\}\\
\\
Explanation:\\
The sequence of instructions can be interpreted as follows:\\
\\
1. Turn around: Move backward 1 step.\\
2. Turn right: Turn 90 degrees clockwise.\\
3. Take 5 steps: Move forward 5 steps.\\
4. Take 2 steps: Move forward 2 steps.\\
\\
The path taken is a straight line with a 90-degree turn at the end, resulting in a total distance of 7 steps. Since the journey ends at the starting point, the answer is "A".\\
\\
\#\#\# Sample 2, 3, 4\\
...

}\\
\hline
Feedbacks&
\tabincell{ p{13cm}}{\\

1. The assistant could have misunderstood the definition of the starting point. In these examples, not all actions are returning the model to the original position, yet the assistant has concluded that they do. It appears the assistant is not correctly interpreting steps that involve turning or moving in directions other than forward. 
\\\\
2. The assistant could be mishandling the concept of directions and movement in a two-dimensional context. For example, it may not be accurately processing how "Take 2 steps backward" or "Take 8 steps left" impacts the end position relative to the starting point. \\\\
3. The assistant could be incorrectly interpreting the phrases "Turn around", "Turn right" and "Always face forward". Its current understanding of these terms might not correctly reflect the changes in direction they imply, leading to inaccurate calculations of the final position. \\
\\
}
\\
\hline
Refined Prompt&
\tabincell{ p{13cm}}{\\
You will be given a series of navigation instructions. In this context, treat 'forward' as movement in the upward direction, 'backward' as downward, 'left' as movement to the left, and 'right' as movement to the right. Based on these instructions, provide a brief explanation of the path followed. Then, check if the journey ends where it started. If the starting point and ending point are the same, give your answer as \{"answer": "A"\}, otherwise, respond with \{"answer": "B"\}.
}
\\\\
\hlineB{3}
\end{tabular}}
\caption{Case of APO prompt updating, obtained with Llama-2-70B-chat.}
\label{tab:APO_case}
\end{table*}


\begin{table*}[h]
\centering
\begin{tabular}{p{5cm} p{8cm}}
\hline
Meta Format Name & Meta Format
\\
\hline
\\
\makecell[l]{
reflection\_format\&\\
prompt\_refinement\_format}&
\makecell[l]{
I have an instruction for a task:\\
"\\
\texttt{[} ins \texttt{]}\\
"\\\\
However, this instruction got the following samples \\
wrong ("\#\# Input" is followed by the input, "\#\# \\
Output" is followed by the follower output. "\#\# \\
Correct Answer" is followed by the correct answer.):\\\\
\texttt{[} error\_examples \texttt{]}\\\\\\
Now, I need you to:\\
1. Read each error sample and carefully analyze why\\
the follower makes mistake. \\
2. Group the causes of errors into several categories.\\
Based on the categories, provide several suggestions \\
for improving the instruction.\\
3. Based on the analysis, please SLIGHTLY revise \\
the instruction to a better one. Wrap the new ins-\\
truction with $<$START$>$ and $<$END$>$.\\
}
\\
\hline
\\
error\_example&
\makecell[l]{
\#\#\# Sample \texttt{[}index\texttt{]}\\
\#\# Input\\
\texttt{[} input \texttt{]}\\
\#\# Correct Answer\\
\texttt{[} answer \texttt{]} \\
\#\# Output\\
\texttt{[}llm\_output \texttt{]}
}\\
\\
\hline

\end{tabular}
\caption{The meta prompt formats used for APO-Sum.}
\label{tab:APO_sum_meta}
\end{table*}

\begin{table*}[h]
\centering
\scalebox{0.75}{
\begin{tabular}{c c}
\hlineB{3}

\multicolumn{2}{c}{Case of APO-Sum}
\\
\hline
\\
Original Prompt&
\tabincell{ p{16cm}}{
You are given a sequence of navigation instructions. Please interpret these instructions in a two-dimensional context (considering left, right, forward, and backward movements). Based on these instructions, provide a concise explanation of the path taken. Then, determine if the journey ends at the starting point. If yes, respond with "A"; if no, respond with "B". Your response should be in the following format: \{"answer": "A"\} or \{"answer": "B"\}.
}\\
\\
\hline\\
Error Examples&
\tabincell{ p{16cm}}{
\#\#\# Sample 1\\
\#\# Input\\
Turn around. Turn right. Take 5 steps. Take 2 steps.\\
\#\# Correct Answer\\
\{"answer": "B"\}\\
\#\# Output\\
\{"answer": "A"\}\\
\\
Explanation:\\
The sequence of instructions can be interpreted as follows:\\
\\
1. Turn around: Move backward 1 step.\\
2. Turn right: Turn 90 degrees clockwise.\\
3. Take 5 steps: Move forward 5 steps.\\
4. Take 2 steps: Move forward 2 steps.\\
\\
The path taken is a straight line with a 90-degree turn at the end, resulting in a total distance of 7 steps. Since the journey ends at the starting point, the answer is "A".\\
\\
\#\#\# Sample 2, 3, 4\\
...

}\\
\hline
\makecell[l]{Feedbacks \& \\Refined Prompts}&
\tabincell{ p{16cm}}{\\
1. Error Analysis:\\
- In all samples, the assistant misunderstands the navigation instructions. It is not properly interpreting "left", "right", "forward" and "backward" in the context of a two-dimensional grid. The assistant appears to consider all the movements cumulatively, rather than as individual movements in different directions. This leads to incorrect path interpretation and wrong determination of ending point. For instance, in "Take 1 step. Take 7 steps. Take 1 step", it considers all steps in the same direction, hence wrongly concluding that the journey ends at the starting point.\\
\\
2. Categories of Errors:\\
- Misinterpretation of Navigation Instructions: The assistant does not correctly interpret the directions (right, left, etc.). \\
- Incorrect Calculations: The assistant doesn’t correctly calculate the total steps taken in each direction.\\
- Misinterpretation of Directional Changes: The assistant does not correctly acknowledge the change in direction. For instance, in "Turn around. Turn right. Take 5 steps. Take 2 steps" it doesn't comprehend that "Turn around" and "Turn right" change the direction of the movement.\\
\\
Suggestions:\\
- The instructions should highlight that the assistant needs to consider each movement separately in the respective direction.\\
- The instruction should make clear that, "Turn right" or "Turn left" means changing the direction by 90 degrees right or left.\\
- It should be made clear that "Turn around" means a 180-degree change in direction (meaning to move in the opposite direction).\\
\\
3. Revised Instruction:\\
\\
$<$START$>$ You are provided with a series of navigation instructions. Interpret these instructions in a two-dimensional grid, considering the movement in four directions: left, right, forward, and backward. Each direction should be treated distinctly, and each change in direction should be accounted for. Consider "Turn right" or "Turn left" as a 90 degrees shift in the respective direction, and "Turn around" as a 180-degree shift (direct opposite direction). Based on these instructions, break down the path taken. Determine if the journey ends at the starting point. If it does, respond with \{"answer": "A"\}, and if it doesn't, respond with \{"answer": "B"\}.$<$END$>$\\
\\
}
\\
\hlineB{3}
\end{tabular}}
\caption{Case of APO-Sum prompt updating, obtained with Llama-2-70B-chat.}
\label{tab:APO_sum_case}
\end{table*}


\begin{table*}[h]
\centering
\small
\scalebox{1.0}{
\begin{tabular}{p{5cm} p{8cm}}

\hline
Meta Format Name & Meta Format
\\
\hline
\\
reflection\_format&
\tabincell{p{8cm}}{
I'm writing prompts for a language model designed \\
for a task.\\\\
My current prompt is: \\
\texttt{[} ins \texttt{]}\\\\
But this prompt gets the following examples wrong:\\
\texttt{[} error\_examples \texttt{]} \\\\
For each wrong example, carefully examine each question and wrong answer step by step, provide comprehensive and different reasons why the prompt leads to the wrong answer. At last, based on all these reasons,summarize and list all the aspects that can improve
the prompt.
}
\\
\hline
\\
error\_example\_format&
\tabincell{p{8cm}}{
\texttt{[} index \texttt{]}\\
The model's input is:\\
\texttt{[} input \texttt{]}\\\\
The model's response is:\\
\texttt{[} output \texttt{]}\\
The correct label is: \texttt{[} answer \texttt{]}\\
The model's prediction is \texttt{[} llm\_output \texttt{]}
}\\
\\
\hline 
\\
prompt\_refinement\_format&
\tabincell{p{8cm}}{
I'm writing prompts for a language model designed
for a task.\\\\
My current prompt is:\\
\texttt{[} ins \texttt{]}\\\\
But this prompt gets the following examples wrong:\\
\texttt{[} error\_examples \texttt{]}\\\\
Based on these errors, the problems with this prompt and 
the reasons are:\\
\texttt{[} feedbacks \texttt{]}\\\\
Based on the above information, please write 
2 new prompts
following these guidelines:\\
1. The new prompts should solve the current prompt's
problems.\\
2. The new prompts should consider the list of prompts
and evolve based on the current prompt.\\
3. Each new prompt should be wrapped with $<$START$>$
and $<$END$>$. \\ \\
The new prompts are:
}
\\
\hline
\end{tabular}}
\caption{The meta prompt formats used for PromptAgent.}
\label{tab:meta_agent}
\end{table*}

\begin{table*}[h]
\centering
\scalebox{0.65}{
\begin{tabular}{p{3cm} p{18cm}}
\hlineB{3}

\multicolumn{2}{c}{Case of PromptAgent}
\\
\hline
\\
Original Prompt&
\tabincell{ p{18cm}}{
You are given a sequence of navigation instructions. Please interpret these instructions in a two-dimensional context (considering left, right, forward, and backward movements). Based on these instructions, provide a concise explanation of the path taken. Then, determine if the journey ends at the starting point. If yes, respond with "A"; if no, respond with "B". Your response should be in the following format: \{"answer": "A"\} or \{"answer": "B"\}.
}\\
\\
\hline\\
Error Examples&
\tabincell{ p{18cm}}{
1\\
The model's input is:\\
Turn around. Turn right. Take 5 steps. Take 2 steps.\\
Choices:\\
A. True\\
B. False\\
\\
The model's response is:\\
\{"answer": "A"\}\\
\\
Explanation:\\
The sequence of instructions can be interpreted as follows:\\
\\
1. Turn around: Move backward 1 step.\\
2. Turn right: Turn 90 degrees clockwise.\\
3. Take 5 steps: Move forward 5 steps.\\
4. Take 2 steps: Move forward 2 steps.\\
\\
The path taken is a straight line with a 90-degree turn at the end, resulting in a total distance of 7 steps. Since the journey ends at the starting point, the answer is "A".\\
\\
The correct label is: B\\
The model's prediction is A\\
\\
2, 3, 4\\
...

}\\
\hline
Feedbacks&
\tabincell{ p{18cm}}{\\
Example 1:\\
In this example, the model didn't calculate the direction after the turn correctly. After turning around and turning right, the model should then go off at a 90 degree angle from the starting point, not back along the same line. We can infer that the model might not have a clear understanding of how to handle 'turn around' and 'turn right', or how to calculate steps after changing directions.\\
\\
Example 2:\\
The model wrongly assessed that taking 1 step, then 7 steps, and finally 1 step would lead back to the starting point. It seems that the model may have misinterpreted the instructions, thinking that the one step at the end was in the opposite direction, therefore negating one of the previous steps.\\
\\
Example 3:\\
The model seemed to misunderstand the 'Take 2 steps backward' instruction, treating it as if the model was moving two steps forward. This may be because the prompt didn't clearly define how to handle 'backward' movements.\\
\\
Example 4:\\
In this case, the model did not consider that the direction to the left and the direction to the right would offset each other. It seemed to treat "left" and "right" as the same direction, rather than opposite ones which would neutralize each other's effect. \\
\\
Improvement Aspects for the Prompt:\\
1. Make it clear that the model should take into account direction changes when interpreting the navigation instructions. 'Left', 'right', 'forward' and 'backwards' should be treated as separate and distinct directions.
2. Clarify how 'turn around' and 'turn right' or 'turn left' should affect the current direction of movement.\\
3. Define more clearly how 'backward' and 'forward' movements should be interpreted in the context of the current direction faced.\\
4. Make sure the model understands that the instruction to take a step in a certain direction means it should move that way from its current position and orientation, not from the original starting point.\\
5. The model needs to understand that the steps in different directions don't necessarily cancel each other out. It needs to keep a proper track of the path it has followed.\\\\
}
\\
\hline
Refined Prompt&
\tabincell{ p{18cm}}{\\
You are provided with a series of navigational instructions which involve turning and moving forward, backward, left, or right. Your task is to interpret these instructions as if you are navigating a grid, where each step moves you one unit on the grid. Keep in mind that 'turning around' rotates your direction by 180 degrees and 'turning right or left' rotates your direction by 90 degrees. Aside, steps taken 'backward' and 'forward' are relative to your current orientation. Also, moving 'left' or 'right' does not change your facing direction but moves you perpendicularly. Consider all these while interpreting the instructions and tracking your path. At the end of the path, you need to determine if you have returned to your original starting point or not. If yes, respond with \{"answer": "A"\}, if no, respond with \{"answer": "B"\}.
}
\\\\
\hlineB{3}
\end{tabular}}
\caption{Case of PromptAgent prompt updating, obtained with Llama-2-70B-chat.}
\label{tab:agent_case}
\end{table*}


\begin{table*}[h]
\centering
\begin{tabular}{p{5cm} p{8cm}}
\hline
Meta Format Name & Meta Format
\\
\hline
\\
prompt\_refinement\_format&
\makecell[l]{
Generate a variation of the following instruction wh-\\
ile keeping the semantic meaning .\\
Input: \\
\texttt{[} ins \texttt{]}\\
Output:\\
}
\\
\hline
\end{tabular}
\caption{The meta prompt formats used for APE.}
\label{tab:meta_APE}
\end{table*}

\begin{table*}[h]
\centering
\scalebox{0.85}{
\begin{tabular}{p{3cm} p{13cm}}
\hlineB{3}

\multicolumn{2}{c}{Case of APE}
\\
\hline
\\
\makecell[l]{Meta Prompt \\for Updating}&
\tabincell{ p{13cm}}{
Generate a variation of the following instruction while keeping the semantic meaning.\\
Input: \\
You are given a sequence of navigation instructions. Please interpret these instructions in a two-dimensional context (considering left, right, forward, and backward movements). Based on these instructions, provide a concise explanation of the path taken. Then, determine if the journey ends at the starting point. If yes, respond with "A"; if no, respond with "B". Your response should be in the following format: \{"answer": "A"\} or \{"answer": "B"\}.\\
\\
Output:\\
}\\
\\
\hline\\

Updated Prompt&
\tabincell{ p{13cm}}{
You are presented with a series of navigational directions. Please read and interpret these directions within a two-dimensional framework (which includes going left, right, forward and backward). Based on the instructions you receive, produce a simple summary of the travelled route. Then, ascertain if the final location is the same as the starting position. If it is, your answer should be "A"; if it isn't, your answer should be "B". Your answer should be structured as follows: \{"answer": "A"\} or \{"answer": "B"\}.
}\\
\\
\hlineB{3}
\end{tabular}}
\caption{Case of APE prompt updating.}
\label{tab:APE_case}
\end{table*}


\begin{table*}[h]
\centering
\begin{tabular}{p{5cm} p{8.5cm}}
\hline
Meta Format Name & Meta Format
\\
\hline

prompt\_refinement\_format&
\makecell[l]{\\
Your task is to generate the instruction $<$INS$>$. \\
Below are some previous instructions with their scores.\\
The score ranges from 0 to 100. \\
\texttt{[} prompts\_scores \texttt{]}\\\\
Below are some problems.\\
\#\#Problems:\\\\
\texttt{[} error\_examples \texttt{]}\\\\
Generate an instruction that is different from all the \\
instructions $<$INS$>$ above, and has a higher score \\
than all the instructions $<$INS$>$ above. The instruction \\
should begin with $<$INS$>$ and end with $<$/INS$>$. The \\
instruction hould be concise, effective  and general-\\
ly applicable to all problems above.
}

\\

\hline
\\
prompts\_scores\_format&
\makecell[l]{
text:\\
\texttt{[} ins \texttt{]}\\
score:\\
\texttt{[} score \texttt{]}
}\\
\\
\hline\\
error\_examples\_format&
\makecell[l]{
\#\#\# Sample \texttt{[} index \texttt{]}\\
\#\# Input\\
\texttt{[} input \texttt{]}\\
\#\# Output\\
\texttt{[}output\texttt{]}
}\\
\\
\hline

\end{tabular}
\caption{The meta prompt formats used for OPRO.}
\label{tab:meta_OPRO}
\end{table*}

\begin{table*}[h]
\centering
\scalebox{0.7}{
\begin{tabular}{p{3cm} p{15cm}}
\hlineB{3}

\multicolumn{2}{c}{Case of OPRO}
\\

\hline\\
\makecell[l]{History Scores}&
\tabincell{ p{15cm}}{
Your task is to generate the instruction $<$INS$>$. Below are some previous instructions with their scores. The score ranges from 0 to 100. \\\\
text:\\
Given a series of navigation instructions, determine whether one would end up back at the starting point after following the instructions. The instructions will include steps and directions. You need to provide the answer as a JSON object, with the key "answer" and the value as one of the choices provided (either "A" for True or "B" for False).\\
score:\\
44\\\\
text: \\
Determine whether one would end up back at the starting point after following a series of navigation instructions. Read the given instructions and choose either "A. True" or "B. False". Provide the answer in the following format: \{"answer": "A"\} or \{"answer": "B"\}.\\
score:\\
52\\\\....\\\\

text:\\
Read the given navigation instructions and determine if one would end up back at the starting point after following them. Choose the correct option from the given choices (A for True, B for False) and provide the answer in the following format: \{"answer": "A"\} or \{"answer": "B"\}.\\
score:\\
62\\ \\
text: \\
Given a series of navigation instructions, determine whether one would end up back at the starting point. Write the answer in the format \{"answer": "A"\} for True or \{"answer": "B"\} for False. Make sure to read and follow the given navigation instructions to reach a conclusion.\\
score:\\
68\\
}\\
\hline\\
Error Examples&
\tabincell{p{15cm}}{
\#\#\# Sample 1\\
$<$INS$>$\\

\#\# Input\\
Turn around. Take 4 steps. Take 7 steps. Turn around. Take 7 steps.\\
Choices:\\
A. True\\
B. False\\
\#\# Output\\
 \{"answer": "A"\}\\\\
 Explanation:\\\\
 After following the navigation instructions, one would end up back at the starting point. Here's a step-by-step breakdown:\\\\
 1. Turn around: Faces the opposite direction.\\
 2. Take 4 steps: Moves 4 steps away from the starting point.\\
 3. Take 7 steps: Moves 7 steps away from the current position.\\
 4. Turn around: Faces the opposite direction again.\\
 5. Take 7 steps: Moves 7 steps towards the starting point.\\\\
 Since the instructions include turning around twice and taking 7 steps in both directions, the net movement is zero, meaning the final position is the same as the starting point. Therefore, the answer is "A. True".\\\\\\
\#\#\# Sample 2,3,4\\
...
}\\
\hline\\
Updated Prompt&
\tabincell{ p{15cm}}{
Analyze the given series of navigation instructions and ascertain whether they would guide you back to the starting point. The instructions will include directions and steps. Choose your answer from two options: A. True (back to the starting point) or B. False (not back to the starting point). Express your response as a JSON object in the following manner: \{"answer": "A"\} or \{"answer": "B"\}.}\\
\\
\hlineB{3}
\end{tabular}}
\caption{Case of OPRO prompt updating.}
\label{tab:OPRO_case}
\end{table*}


\begin{table*}[h]
\centering
\small
\scalebox{0.79}{
\begin{tabular}{|c|}
\hline
\tabincell{p{18cm}}{
\\You are presented with several pieces of feedback and a number of pre-existing clusters.\\\\
For each piece of feedback, analyze it thoroughly and consider:\\\\
Can the issue identified in the feedback be categorized into ONE unique pre-existing cluster?\\
Please format your evaluation as follows:\\\\
\ \ \ \ \ \#Feedback ID: [Insert your analysis here]\\
\ \ \ \ \ [Identify key problem outlined in feedback i]\\
\ \ \ \ \ [Discuss relation between feedback i \& cluster 1,2,...,n and provide your explanation]\\\\
    
\ \ \ \ \ - If a piece of feedback can be categorized into a UNIQUE pre-existing cluster, indicate this by using the \textless Cluster \textgreater and \textless /Cluster \textgreater tags. It's crucial to note that one piece of feedback can only belong to one cluster.\\
\ \ \ \ \ - If it CANNOT, create a new cluster for this conducted feedback. Indicate the creation of a new cluster using the \textless New Cluster\textgreater [description of the cluster] \textless /New Cluster\textgreater tags. Note: the description of the cluster must be clear and explicit.\\\\\\

\textcolor[RGB]{6,128,67}{
\#\#\#Examples}\\\\
\textcolor[RGB]{6,128,67}{\#\#Existing Clusters}\\\\

\textcolor[RGB]{6,128,67}{Cluster1: The prompt is not explicit enough about how to handle sentences that include numbers}\\
\textcolor[RGB]{6,128,67}{Cluster2: the prompt does not specify whether 'a' and 'an' should be counted as '1'}\\\\

\textcolor[RGB]{6,128,67}{\#\#Feedback}\\\\

\textcolor[RGB]{6,128,67}{Feedback 1}\\
\textcolor[RGB]{6,128,67}{Misinterpretation of the Task: The prompt may not accurately convey the objective to the model. The model should count individual objects, taking into account multiple instances of the same object. However, the prompt could potentially be misunderstood as counting types of objects, as opposed to individual instances resulting in an inaccurate count. This could be seen in all examples, where the model appears to count unique objects rather than the total quantity.}\\
\textcolor[RGB]{6,128,67}{Feedback 2}\\
\textcolor[RGB]{6,128,67}{A second possible issue could be the absence of clear instructions on how to count objects linked with the words 'a' and 'an'. This ambiguity may cause confusion, especially with the unclear statement "The objects may be of different types and some may be specifically mentioned in multiples (like 'two chickens')". This lack of instruction may cause the model to miscount objects associated with 'a' and 'an', leading to incorrect results.}\\\\

\textcolor[RGB]{6,128,67}{Solution:}\\\\

\textcolor[RGB]{6,128,67}{\#Feedback1: Misinterpretation of Task as Counting Types of Objects as opposed to Counting Individual Objects}\\\\

\textcolor[RGB]{6,128,67}{Feedback 1 details a different issue where the model misunderstanding the task - it counts types of objects, instead of individual objects, leading to an inaccurate count.}\\\\

\textcolor[RGB]{6,128,67}{Feedback 1 \& Cluster 1: While Feedback 2 does touch on a counting issue, it doesn’t fall within the scope of Cluster 1 since it's not about sentences involving numbers. }\\\\

\textcolor[RGB]{6,128,67}{Feedback 1 \& Cluster 2: Similarly, this feedback does not discuss 'a' and 'an', thus it doesn't align with Cluster 2.}\\\\

\textcolor[RGB]{6,128,67}{Given that Feedback 1 identifies a separate issue that isn't currently covered by either cluster, it is necessary to create a new cluster: \textless New Cluster\textgreater Cluster3: The prompt leads to misinterpretation of the task as Counting Types of Objects as opposed to Counting Individual Objects\textless /New Cluster\textgreater .}\\\\\\

\textcolor[RGB]{6,128,67}{\#Feedback2: Unclear instructions on how to count objects linked with the words 'a' and 'an'}\\\\

\textcolor[RGB]{6,128,67}{Feedback 2 revolves around the problem of counting objects linked with 'a' and 'an'. This feedback states that there are no clear instructions on these instances which likely leads to the model miscounting objects.}\\\\

\textcolor[RGB]{6,128,67}{Feedback 2 \& Cluster 1: This feedback doesn't relate to Cluster 1 as it doesn't touch upon the issue of handling sentences with numbers. }\\\\

\textcolor[RGB]{6,128,67}{Feedback 2 \& Cluster 2: This feedback correlates with the problem outlined in Cluster 2, specifically the confusion around counting instances of 'a' and 'an'. Hence, this feedback should be organized under \textless Cluster\textgreater Cluster2: the prompt does not specify whether 'a' and 'an' should be counted as '1'\textless /Cluster\textgreater . }\\\\

\textcolor[RGB]{6,128,67}{Feedback 2 \& Cluster 3: This feedback doesn't relate to Cluster 3 as it doesn't discuss the misunderstanding of the prompt's main objective.}

\\\\
Now, complete the following task as instructed:\\\\
\#\#Existing Clusters\\\\
\textcolor[RGB]{26,103,240}{[all the exiting cluster till now]}
\\\\
\#\#Feedback\\\\
\textcolor[RGB]{26,103,240}{[a batch of feedbacks to be clustered]}\\\\\\
Solution:
}
\\
\hline
\end{tabular}}
\caption{Prompt for GPT-4-based feedback clustering.}
\label{tab:cluster_prompt}
\end{table*}


\begin{algorithm}
\caption{GPT-4-based Feedback Clustering Algorithm}
\small
\begin{algorithmic}[1]\label{alg:cluster}

\STATE \textbf{Input:} 
\STATE $FD_{1..n}$       \ \ \ \ \ \ \ \  \ \ \   \ \ // Feedback lists for method 1,2,...,n  
\STATE $batch\_size$    \  \ \ \ \ \  \   // The number of feedback  in  a batch 
\STATE Template \ $\mathcal{T}$    \  \ \ \ \ \ // The prompt template for clustering \\

\STATE \textbf{Initialize:} 
\STATE $D_f  \leftarrow \{ \ \}$    \ \ \ \ \ \  \ \  \ \ \ \ \ \ \ \ \  \ \ \  // Dict for all feedbacks \\
\STATE $C_{exist}  \leftarrow [\ ]$   \ \ \ \ \ \ \ \ \ \ \ \ \ \ \ \ \ // List for existing clusters
\STATE $C_{result} \leftarrow \{\ \}$  \ \ \ \ \ \  \ \ \ \ \ \ \ \  // Dict for clustering results

\FOR {method\_index $ \leftarrow  0,1...,n-1$}
  \FOR {each feedback $\in FD_{method\_index}$}
    \STATE $D_f \leftarrow D_f + $ (feedback,method\_index)
  \ENDFOR
\ENDFOR

\STATE batch\_num $ \leftarrow \|D_f\| / batch\_size$
\FOR {i in $ 0,... ,  batch\_num-1$}
  \STATE batch\_feedback $ \leftarrow D_f [i*batch\_size: i*(batch\_size+1)]$
  \STATE query $ \leftarrow \mathcal{T}($batch\_feedback$,C_{exist})$ 
  \STATE cluster\_result $ \leftarrow $ GPT-4(query) \ \ \ // Query GPT-4
  \FOR {$(f_i,c_i) \in$ zip (batch\_feedback,cluster\_result)}
     \IF{$c_i$ $\notin C_{exist}$} 
        \STATE $C_{exist} \leftarrow c_i$
     \ENDIF
     \STATE $C_{result} \leftarrow (f_i,c_i)$
  \ENDFOR
\ENDFOR \\
\STATE \textbf{Return} $C_{result}$
\end{algorithmic}
\end{algorithm}

\begin{table*}[h]
\centering
\small
\scalebox{1}{
\begin{tabular}{cc}
\hlineB{3}
\\
\centering Cluster1& 
\tabincell{p{12cm}}{
\textcolor[RGB]{26,103,240}{the prompt does not specify whether 'a' and 'an' should be counted as '1' }\\\\
The second reason might be that the model does not handle well the concept of singularity and plurality in the sentence. When there's only one item ('a' or 'an' used), it may overlook it because it doesn't interpret 'a' or 'an' as representing 'one'. 
\\\\
The first reason could be that the counting mechanism is misunderstanding the use of the word "a" or "an" before a singular noun. Instead of correctly counting it as "1", it might be counting it as "2". This is evident in all the given examples where the response is always one count more than the correct answer.}\\\\
\hline
\\
\centering Cluster2&
\tabincell{p{12cm}}{
\textcolor[RGB]{26,103,240}{Difficulty in Understanding and Counting Items When Numbers are Mentioned as Words}
\\\\
Lastly, there could be an issue with the interpretation of the numeric words (like "three", "four", etc.). The prompt might not be clear enough in instructing that the numbers before an object need to be multiplied with the object to get the correct count. Hence, it might count 'four fridges' as one fridge, leading to an incorrect total.\\\\
Lastly, the prompt fails to emphasize the importance of understanding the semantics of the input text. This includes numerical meaning and its role in providing the correct answer. For instance, when the text says "three trumpets", the model should understand that it needs to count this as three items, not one. Thus, natural language processing and comprehension is a crucial aspect that the prompt should stress more on.}\\\\
\hline
\\
\centering Cluster3&
\tabincell{p{12cm}}{
\textcolor[RGB]{26,103,240}{Misinterpretation Due to Complex Sentence Structures or Presence of Extra Words}\\\\
Lastly, the problem could lie with the model's comprehension of complex sentences. Given the structure of the sentences in the examples, it is possible that the model is not correctly parsing the sentences and is not identifying all the objects or their quantities. This could be due to issues with how it processes conjunctions, commas, or the specific syntax used, resulting in a lower than accurate count of objects.\\\\The model might be having trouble understanding complex sentences: The model could struggle with longer sentences, especially those that include a mix of single and multiple items, leading to inaccurate counts. It might be missing an item when many are listed in one sentence, misinterpreting the grammar, or failing to correctly parse sentences with more intricate structures.}\\\\
\hline
\\
\centering Cluster4&
\tabincell{p{12cm}}{
\textcolor[RGB]{26,103,240}{The prompt is unclear on how to handle plurals in the sentence}\\\\
Reason 3: The system might be inaccurately processing plural items. For example, if the prompt mentions "five chairs", the system needs to understand that it's 5 individual items, not just a single entity. If the system doesn't have an understanding of singular and plural nouns or isn't implemented to handle them, it will likely count them incorrectly.\\\\The prompt may be unclear on how to handle plurals in the sentence. It's not specified whether the model should consider plurality to mean more than one object, which could affect the count. For example, in the first case, 'trumpets' could be interpreted as one object instead of two, leading to a miscount.}\\\\
\hlineB{3}
\end{tabular}
}
\caption{Cluster descriptions and example feedbacks for the feedback clustering experiment (1/3).}
\label{tab:cluster_1}
\end{table*}

\begin{table*}[h]
\centering
\small
\scalebox{1}{
\begin{tabular}{cc}
\hlineB{3}
\\
\centering Cluster5&
\tabincell{p{12cm}}{
\textcolor[RGB]{26,103,240}{The prompt is unclear on how to handle situations where the same object is mentioned multiple times}
\\\\
Lastly, the issue could lie in the instruction about objects that are usually mentioned only once. The current wording could lead to the assumption that if an object is mentioned more than once, the additional mentions should not be counted. This could be why the model is returning a lower count than expected, especially in the instances where the same object (e.g., a flute) is mentioned more than once in the list.\\\\The third reason could be that the prompt is not specific enough with its instructions regarding how to handle multiple instances of the same item. For example, in the sentence "I have a cat, a cat, and a cat", it's unclear from the prompt whether this should be counted as three items or one item. Confusion from this lack of specification may be causing the model to produce incorrect counts.}\\\\
\hline
\\
\centering Cluster6&
\tabincell{p{12cm}}{
\textcolor[RGB]{26,103,240}{The prompt is unclear about how to correctly parse the items when they are grouped together with commas, conjunctions, punctuation or "and".}\\\\
Reason 2: There might be an issue with how the system is differentiating and identifying different items in a sentence, especially if they are separated by commas or linked by 'and'. If it is not isolating each item effectively, the count it provides will be off. The system should comprehend that each time an item is mentioned, whether it's a fruit, a piece of furniture, a musical instrument, or an animal, it's a separate entity that needs to be counted.\\\\The third reason could be due to incorrect processing of the sentence structure. If the sentence is parsed incorrectly, whether due to punctuation, conjunctions, or the use of commas to list items, this could impact the ability to correctly identify and count the individual units mentioned in the sentence. Incorrect parsing could lead to missing or miscounting items, resulting in an inaccurate total count.}\\\\
\hline
\\
\centering Cluster7&
\tabincell{p{12cm}}{
\textcolor[RGB]{26,103,240}{The prompt does not clear about how to handle compound items or certain phrases such as "a head of broccoli".}\\\\
The prompt does not explicitly state how to deal with compound objects like "lettuce heads," "stalk of celery," or 'head of broccoli'. The model may be incorrectly counting these as multiple objects when they should be counted as one.\\\\The prompt might be confused by complex object names: In the examples given, complex names like "a head of broccoli" or "a stalk of celery" could be causing confusion for the model. It might not be recognizing these as single objects because of the additional describing words (head, stalk) and so it might be undercounting them. The prompt's instructions need to include information on how to handle these unique cases.}\\\\
\hline
\\
\centering Cluster8&
\tabincell{p{12cm}}{
\textcolor[RGB]{26,103,240}{The prompt leads to misinterpretation of the task as Counting Types of Objects as opposed to Counting Individual Objects }
\\\\
The prompt might not be explicit enough in stating that the model's answer should be a sum of all individual items, not just a count of the unique types of items. This could lead the model to just count different types of items rather than adding up their quantities.\\\\Reason 1: The current prompt can be misinterpreted, as it might let the AI system think that it needs to count the unique objects instead of the total number of objects. The prompt says to count the total number of objects but doesn't specify that the count should include multiple instances of the same object, which might lead to incorrect outputs.}\\\\
\hlineB{3}
\end{tabular}
}
\caption{Cluster descriptions and example feedbacks for the feedback clustering experiment (2/3).}
\label{tab:cluster_2}
\end{table*}

\begin{table*}[h]
\centering
\small
\scalebox{0.9}{
\begin{tabular}{cc}

\hlineB{3}
\\
\centering Cluster9&
\tabincell{p{13.2cm}}{
\textcolor[RGB]{26,103,240}{The prompt does not specify the format in which the output should be provided}\\\\
Reason 2: The prompt doesn't specify the format in which the output should be provided beyond the fact that it should be a dictionary with the key "answer". This could be causing the model to interpret the output format in different ways, as shown in Sample 3 where the model lists out the individual counts for each item instead of providing a total sum.\\\\A third potential issue could be the way the prompt describes how to output the answer. The term 'structured format' might be ambiguous and cause confusion about the format in which the final count should be presented. For example, it can be unclear whether the count should be a number or a string within the braces. To overcome this, the instructions need to clarify the format of the answer more explicitly.}\\\\
\hline
\\
\centering Cluster10&
\tabincell{p{13.2cm}}{
\textcolor[RGB]{26,103,240}{The task prompt is not clear about what constitutes an "object"}\\\\
Ambiguity in defining "objects": The task prompt is not clear about what constitutes an "object". Is it any noun mentioned in the text or only specific types of items (for example, tangible items)? This might have led to the model counting extra items, such as intangible items or abstract nouns, in the input text.
\\\\The third reason could be a misunderstanding of what constitutes an "item". The prompt might not be clearly defining what an "item" is, which could cause the model to misinterpret and incorrectly count the given items in the sentence."}\\\\
\hline
\\
\centering Cluster11&
\tabincell{p{13.2cm}}{
\textcolor[RGB]{26,103,240}{The model may lack understanding of the counting process, which might due to inadequate training}\\\\
Secondly, the model may not fully understand the concept of 'counting' as it applies to this task. Even though the prompt specifies to count individual items and even provides an example, the model seems to be making a consistent error of undercounting. This could be due to a lack of understanding of the counting process or a computational error.\\\\The third reason could be due to the model not being trained or programmed to understand the concept of counting items from language input. The model might not be competent in tasks that require more explicit numerical logic or counting, hence leading to inaccurate answers.}\\\\
\hline
\\
\centering Cluster12&
\tabincell{p{13.2cm}}{
\textcolor[RGB]{26,103,240}{The model might mistakenly group certain items together as a group or category and count them as one instead of as individual items}\\\\
The first reason could be the lack of clarity on how to count objects that are part of the same category but are distinctly mentioned separately. For instance, in the first example, different kinds of vegetables are mentioned. It's not clear whether these should be counted as individual objects or if they fall under a broader 'vegetable' category. Similarly, in the second and third examples, different musical instruments are listed, causing the same confusion.\\\\Another issue might be with the way the model is interpreting words like "oven, fridges, chair, lamps". Although each item is different, the model might group these items together as a set or category and count them as one instead of as individual items.}\\\\
\hline
\\
\centering Cluster13&
\tabincell{p{13.2cm}}{
\textcolor[RGB]{26,103,240}{The prompt is ambiguous about how to handle objects traditionally counted in multiples}
\\\\
The second reason could be a misinterpretation of the rule regarding objects that are usually counted in multiples. The prompt suggests garlic cloves are considered as a single item unless a specific quantity is implied or stated. However, the system might be considering "a garlic" as multiple objects instead of one, which seems to be the case in Sample 3 and 4.\\\\The second reason might be related to the AI's understanding of collective nouns versus individual objects. For instance, in the context of the prompt, a collective noun like 'a pair of shoes' is to be considered as one object. However, the AI might not have been correctly trained to count this as one object and is instead counting them as two separate items.}\\\\

\\
\hlineB{3}
\end{tabular}
}
\caption{Cluster descriptions and example feedbacks for the feedback clustering experiment (3/3).}
\label{tab:cluster_3}
\end{table*}


\begin{table*}[h]
\centering
\small
\scalebox{1}{
\begin{tabular}{c c}
\hlineB{3}
Current Prompt & Feedback
\\
\hline
\tabincell{p{7.2cm}}{\\
The task is to identify and calculate the total quantity of items mentioned in the input text. The text will feature a variety of items along with their quantities. When an item does not have a specified quantity, you should count it as one. Remember that the aim is not to count individual names types of items, but to consider the quantities specified for each of them. For instance, "three violins" should be recorded as three items, not one.\textbf{ \textcolor[RGB]{228,27,52}{Particularly pay attention to the phrases like "a" or "an" preceding an item as they imply one item.}} The comprehension of the text semantics is critical to correctly implement the quantity values in your count. The final output should be given in a JSON format, with the key "answer" and the output value as a string.\\\\
}&
\tabincell{p{7.2cm}}{
\textbf{\textcolor[RGB]{228,27,52}{The prompt does not clearly specify that words like "a" or "an" also represent a quantity of one.}} The system might be interpreting these as additional items, which is why it's counting more items than there actually are. (classified in Cluster 1.)
}
\\
\hline
\tabincell{p{7.2cm}}{\\
The task involves counting the total number of items mentioned in a sentence. These items could be anything, from animals to musical instruments to furniture, and may include specified quantities.\textbf{ \textcolor[RGB]{228,27,52}{For example, if the sentence mentions "two chairs", it should be counted as two items.}} Once all the items are counted, the total count should be provided as the answer, formatted as a JSON object with the key "answer". The value of "answer" should be a string of the total count. For example, if the total count is 13, the output should be \{"answer": "13"\}.\\\\
}&\tabincell{p{7.2cm}}{
The first reason could be that the model is not correctly interpreting quantities before items. \textbf{\textcolor[RGB]{228,27,52}{For instance, when the sentence mentions "three trumpets", it should count as three items, not one.}} Misunderstanding this rule could lead to inaccuracies like in the given examples. (classified in Cluster 2.)
}
\\
\hline
\tabincell{p{7.2cm}}{\\
Your task is to count the total number of objects referred to in a given sentence, and return this count. In this task, an object can refer to anything tangible like animals, musical instruments, or household items. If a number is used before an object, that should be considered as the quantity of that particular item. For example, "two hats" should be counted as two separate objects. Phrases like 'a bunch of grapes' or 'a pair of shoes' are to be considered as one single unit. Items should not be broken down into their individual components; \textbf{\textcolor[RGB]{228,27,52}{for example, a 'car' should not be broken down into 'wheels', 'engine' etc.}} After counting all the items, your response should be a JSON object with the key "answer" and the computed total count as a string as the value. For example, if the total number of items counted is 10, your output should look like this: \{"answer": "10"\}.\\\\
}&\tabincell{p{7.2cm}}{
The third reason could be that the model is not able to correctly distinguish between compound objects and individual objects. \textbf{\textcolor[RGB]{228,27,52}{For instance, in examples where a single item like "a car" is mentioned, the model should interpret it as one object.}} However, the model could possibly be misinterpreting these as multiple objects, leading to a higher item count. (classified in Cluster 7.)
}
\\
\hline
\tabincell{p{7.2cm}}{\\
You are tasked with calculating the total count of individual items as per the description in a sentence. Pay attention to the following rules for counting: If you come across a single item such as 'a chair' or 'an orange', count it as one item.\textbf{\textcolor[RGB]{228,27,52}{ If you find a group of objects with a specific number attached, like 'four dogs' or 'seven grapes', each object in the group should be counted as a separate item.}}Hence, 'four dogs' should be treated as four distinct items, not just one. Add up all these counts to get the final total. Your answer should be presented in this manner: \{"answer": "your answer here"\}, substituting 'your answer here' with the total count you calculated as a numerical value. It is essential to understand that every single item, whether it's in a group or not, is to be treated as an individual object and counted towards the total.\\\\
}&\tabincell{p{7.2cm}}{
The first reason could be that \textbf{\textcolor[RGB]{228,27,52}{the prompt might not be clear enough in instructing that each individual item in a group, as well as every single item mentioned, need to be counted separately. }}This could have resulted in the model understanding that it should count groupings of objects (like 'two toasters') as a single item instead of counting each individual object in the group. (classified in Cluster 8.)
}
\\
\hline

\end{tabular}
}
\caption{Cases of repetitive feedbacks for single prompt instances.}
\label{tab:repetitive_case}
\end{table*}

\begin{table*}[h]
\centering
\small
\begin{tabular}{|c|c|c|}
\hline

Prompt
&
\tabincell{p{7cm}}{Given a series of navigation instructions, your task is to determine whether one would end up back at the starting point after following all the instructions. The instructions will use the words left, right, forward, and backward to indicate direction. Each step will be prefixed by a number indicating the number of steps to take in that direction. If a direction is not specified with the step, interpret it as going straight ahead from the initial facing direction. 
A 'turn around' instruction inverts the current facing direction. If the instruction starts with 'Always face forward', all the steps should be taken considering the initial facing direction as forward; left, right and backward steps should be calculated relative to this direction.
After interpreting all the instructions, \textcolor[RGB]{26,103,240}{if one ends up at the starting point, output \{"answer": "A"\}. If one ends up at a different point, output \{"answer": "B"\}.}}
& 
\tabincell{p{7cm}}{\\Given a series of navigation instructions, your task is to determine whether one would end up back at the starting point after following all the instructions. \textbf{\textcolor[RGB]{228,27,52}{First repeat the input without change, then give your answer.} }The instructions will use the words left, right, forward, and backward to indicate direction. Each step will be prefixed by a number indicating the number of steps to take in that direction. If a direction is not specified with the step, interpret it as going straight ahead from the initial facing direction. 

A 'turn around' instruction inverts the current facing direction. If the instruction starts with 'Always face forward', all the steps should be taken considering the initial facing direction as forward; left, right and backward steps should be calculated relative to this direction.

After interpreting all the instructions, \textcolor[RGB]{26,103,240}{if one ends up at the starting point, output \{"answer": "A"\}. If one ends up at a different point, output \{"answer": "B"\}.}\\\\}\\
\hline

Input
&
\tabincell{p{7cm}}{\\
\textbf{\#\# Input} \\
Turn left. Turn around. Take 7 steps. Turn left.Turn right. Turn around. Take 7 steps.
Choices:\\
A. True\\
B. False\\
\textbf{\#\# Output}\\ }
&
\tabincell{p{7cm}}{\\
\textbf{\#\# Input} \\
Turn left. Turn around. Take 7 steps. Turn left.Turn right. Turn around. Take 7 steps.\\
Choices:\\
A. True\\
B. False\\
\textbf{\#\# Output}\\}
\\
\hline
\makecell{
LLM\\
Output
}
&
\tabincell{p{7cm}}{
\textcolor[RGB]{6,128,67}{\{"answer": "A"\}}\\
Explanation:\\
1. Turn left: Face left.\\
2. Turn around: Face right.\\
....\\
so the answer is "A".}
&
\tabincell{p{7cm}}{\\
\textcolor[RGB]{6,128,67}{Input: Turn left. Turn around. Take 7 steps. Turn left. Turn right. Turn around. Take 7 steps.}\\
\textcolor[RGB]{228,27,52}{Output: B. False }\\
Explanation:\\
1. Turn left: Facing left.\\
2. Turn around: Facing right.\\
....\\
Therefore, the answer is B. False.\\\\}
\\
\hline
\end{tabular}
\caption{Case for the instruction-following experiment. This table shows the case of inserting the ``Repeat'' instruction into the prompt for Llama-2-70B-chat. The model failed to strict to the answer format and failed to answer correctly when following the extra guidance.}
\label{tab:following_case_llama_repeat}
\end{table*}


\begin{table*}[h]
\centering
\small
\begin{tabular}{|c|c|c|}
\hline

Prompt
&
\tabincell{p{7cm}}{\\Using a series of navigation instructions, determine if the final location is the same as the starting point. The instructions will provide direction (forward, backward, left, right), and the number of steps to be taken. Treat forward as moving in the same direction one is facing, backward as moving in the opposite direction, left as moving to the left while facing the same direction and right as moving to the right while facing the same direction. Directions are oriented from the starting point. \textcolor[RGB]{26,103,240}{Use the information to decide if one ends up back at the starting point and write \{"answer": "A"\} if true and \{"answer": "B"\} if false.} Instructions without a clear direction cannot be used for this task.\\\\}
& 
\tabincell{p{7cm}}{\\Using a series of navigation instructions, determine if the final location is the same as the starting point. \textbf{\textcolor[RGB]{228,27,52}{Include the keyword "Alright" in your response.}}The instructions will provide direction (forward, backward, left, right), and the number of steps to be taken. Treat forward as moving in the same direction one is facing, backward as moving in the opposite direction, left as moving to the left while facing the same direction and right as moving to the right while facing the same direction. Directions are oriented from the starting point. \textcolor[RGB]{26,103,240}{Use the information to decide if one ends up back at the starting point and write \{"answer": "A"\} if true and \{"answer": "B"\} if false.} Instructions without a clear direction cannot be used for this task.\\\\
}\\
\hline

Input
&
\tabincell{p{7cm}}{\\
\textbf{\#\# Input} \\
Always face forward. Take 6 steps forward. Take 4 steps backward. Take 1 step forward. Take 9 steps forward. Take 7 steps backward. Take 5 steps backward.\\
Choices:\\
A. True\\
B. False\\
\textbf{\#\# Output} }
&
\tabincell{p{7cm}}{\\
\textbf{\#\# Input} \\
Always face forward. Take 6 steps forward. Take 4 steps backward. Take 1 step forward. Take 9 steps forward. Take 7 steps backward. Take 5 steps backward.\\
Choices:\\
A. True\\
B. False\\
\textbf{\#\# Output}}
\\
\hline
\makecell{
LLM\\
Output
}

&
\tabincell{p{7cm}}{\\
\textcolor[RGB]{6,128,67}{\{"answer": "A"\}}\\
Explanation:\\
1. Start facing forward.\\
2. Take 6 steps forward, moving away from the starting point.\\
....\\
the answer is "A. True".\\\\
}
&
\tabincell{p{7cm}}{\\
\textcolor[RGB]{6,128,67}{Alright}, let's navigate!\\
Starting point: Facing forward\\\\
1. Take 6 steps forward: Facing forward\\
2. Take 4 steps backward: Facing backward\\
....\\
\textcolor[RGB]{228,27,52}{so the answer is B. False.}\\\\}
\\
\hline
\end{tabular}
\caption{Case for the instruction-following experiment. This table shows the case of inserting the ``Keyword'' instruction into the prompt for Llama-2-70B-chat. The model failed to strict to the answer format and failed to answer correctly when following the extra guidance.}
\label{tab:following_case_llama_keyword}
\end{table*}


\begin{table*}[h]
\centering
\small
\begin{tabular}{|c|c|c|}
\hline

Prompt
&
\tabincell{p{7cm}}{\\Your task is to interpret a set of navigation instructions and determine the final position relative to the start. The instructions may include steps to the left or right, turning around, or always facing forward. In these instructions, "left" and "right" steps refer to directions relative to the current facing direction, "turning around" implies reversing the current facing direction by 180 degrees, and "always face forward" means the direction remains unchanged throughout. Evaluate the total distance travelled in each direction and determine if the end position matches the start position. \textcolor[RGB]{26,103,240}{If the positions match, respond with \{"answer": "A"\}. If they do not match, respond with \{"answer": "B"\}.}\\\\
}
& 
\tabincell{p{7cm}}{\\Your task is to interpret a set of navigation instructions and determine the final position relative to the start. \textbf{\textcolor[RGB]{228,27,52}{Include the keyword "Alright" in your response.}} The instructions may include steps to the left or right, turning around, or always facing forward. In these instructions, "left" and "right" steps refer to directions relative to the current facing direction, "turning around" implies reversing the current facing direction by 180 degrees, and "always face forward" means the direction remains unchanged throughout. Evaluate the total distance travelled in each direction and determine if the end position matches the start position. \textcolor[RGB]{26,103,240}{If the positions match, respond with \{"answer": "A"\}. If they do not match, respond with \{"answer": "B"\}.}\\\\
}\\
\hline

Input
&
\tabincell{p{7cm}}{\\
\textbf{\#\# Input} \\
Always face forward. Take 6 steps forward. Take 10 steps left. Take 8 steps left. Take 2 steps backward. Take 4 steps backward.\\
Choices:\\
A. True\\
B. False\\
\textbf{\#\# Output} }
&
\tabincell{p{7cm}}{\\
\textbf{\#\# Input} \\
Always face forward. Take 6 steps forward. Take 10 steps left. Take 8 steps left. Take 2 steps backward. Take 4 steps backward.\\
Choices:\\
A. True\\
B. False\\
\textbf{\#\# Output}}
\\
\hline
\makecell{
LLM\\
Output
}

&
\tabincell{c}{\\
\textcolor[RGB]{6,128,67}{\{"answer": "B"\}}\\
}
&
\tabincell{p{7cm}}{\\
\textcolor[RGB]{228,27,52}{Alright}, the final position relative to the start is 2 steps to the left and 6 steps backward. Since the end position does not match the start position, \textcolor[RGB]{228,27,52}{the answer is B.}\\\\
}
\\
\hline
\end{tabular}
\caption{Case for the instruction-following experiment. This table shows the case of inserting the ``Keyword'' instruction into the prompt for GPT-3.5-Turbo. The model failed to strict to the answer format when following the extra guidance.}
\label{tab:following_case_35_keyword}
\end{table*}


\begin{table*}[h]
\centering
\small
\begin{tabular}{|c|c|c|}
\hline

Prompt
&
\tabincell{p{7cm}}{\\You are to analyze a series of navigation commands to decide whether they will guide the navigator back to their starting point. The navigator could be asked to move 'forward', 'backward', 'left', 'right', or to 'turn around'. 'Always face forward' indicates that the navigator's initial orientation is preserved throughout the journey. When told to move 'left' or 'right', the navigator moves laterally in that direction without changing their orientation. 'Turn around' instructs the navigator to face the opposite direction. If the set of instructions brings the navigator back to their starting point, \textcolor[RGB]{26,103,240}{reply with \{"answer": "A"\}. If it does not, reply with \{"answer": "B"\}.}
}
& 
\tabincell{p{7cm}}{\\You are to analyze a series of navigation commands to decide whether they will guide the navigator back to their starting point. \textbf{\textcolor[RGB]{228,27,52}{First repeat the input without change, then give your answer.}} The navigator could be asked to move 'forward', 'backward', 'left', 'right', or to 'turn around'. 'Always face forward' indicates that the navigator's initial orientation is preserved throughout the journey. When told to move 'left' or 'right', the navigator moves laterally in that direction without changing their orientation. 'Turn around' instructs the navigator to face the opposite direction. If the set of instructions brings the navigator back to their starting point, \textcolor[RGB]{26,103,240}{reply with \{"answer": "A"\}. If it does not, reply with \{"answer": "B"\}.}

}
\\
\hline

Input
&
\tabincell{p{7cm}}{\\
\textbf{\#\# Input} \\
Take 3 steps. Take 4 steps. Turn around. Take 1 step. Take 6 steps.
\\
Choices:\\
A. True\\
B. False\\
\textbf{\#\# Output} }
&
\tabincell{p{7cm}}{\\
\textbf{\#\# Input} \\
Take 3 steps. Take 4 steps. Turn around. Take 1 step. Take 6 steps.
\\
Choices:\\
A. True\\
B. False\\
\textbf{\#\# Output}}
\\
\hline
\makecell{
LLM\\
Output
}

&
\tabincell{c}{\\
\textcolor[RGB]{6,128,67}{\{"answer": "A"\}}\\
}
&
\tabincell{p{7cm}}{\\
\textcolor[RGB]{228,27,52}{Take 3 steps. Take 4 steps. Turn around. Take 1 step. Take 6 steps.}\\
\textcolor[RGB]{228,27,52}{\{"answer": "B"\}}\\\\
}
\\
\hline
\end{tabular}
\caption{Case for the instruction-following experiment. This table shows the case of inserting the ``Repeat'' instruction into the prompt for GPT-3.5-Turbo. The model correctly answer when following the extra guidance.}
\label{tab:following_case_35_repeat}
\end{table*}

\begin{table*}[h]
\centering
\small
\scalebox{0.81}{
\begin{tabular}{|c|}
\hline
\tabincell{p{18cm}}{\\Here is a new step-by-step instruction based on the feedback and error analysis:\\
\\
1.  Thoroughly read the given input sentence.\\
  \\
2.  Identify the category of objects you are asked to count. It could be animals, musical instruments, or other categories. \\
    \\
3.  Start counting the objects of that category mentioned in the sentence. Make sure to multiply the count of the object by the number mentioned with it, if any (like 'two chickens' means two objects).\\
    \\
4.  Keep a track of your counting process. \\
    \\
5.  Once you have counted all the objects, sum up all the individual counts and those specifically mentioned in multiples.\\
    \\
6.  Go through the sentence one more time to ensure you haven't missed any objects.\\
    \\
7.  Write down the total count as a string in the format: {"answer": "number"} where 'number' is your calculated total count of objects.\\
\\
\#\#\#\# Examples\\
Here are some examples to help you thinking. \\
\\
\#\# Input\\
I have a garlic, a potato, a trombone, a yam, a lettuce head, a drum, a head of broccoli, a flute, a stalk of celery, a cauliflower, and a cabbage. How many vegetables do I have?\\
\#\# Output\\
\\
Step 1: \\
The given input sentence is 'I have a garlic, a potato, a trombone, a yam, a lettuce head, a drum, a head of broccoli, a flute, a stalk of celery, a cauliflower, and a cabbage. How many vegetables do I have?'\\
\\
Step 2:\\
The category of objects to be counted is vegetables.\\
\\
Step 3 \& 4:\\
Let's begin counting: \\
\\
- Garlic, which is a vegetable, so count is 1.\\
- Potato, this is also a vegetable, so adding this, the count goes to 2.\\
- Trombone, this is a musical instrument, not a vegetable. So, this is not added to the count.\\
- Yam, this is a vegetable, so the count is now 3.\\
- Lettuce head, this is a vegetable as well, so the count is now 4.\\
- Drum, this is a musical instrument, not a vegetable. So, this is not added to the count.\\
- Head of broccoli, this is a vegetable, so the count is now 5.\\
- Flute, this is a musical instrument, not a vegetable. So, this is not added to the count.\\
- Stalk of celery, this is a vegetable, so the count is now 6.\\
- Cauliflower, this is a vegetable as well, so the count is now 7.\\
- Cabbage, this too is a vegetable, so the count is now 8.\\
\\
Step 5:\\
Sum up all the individual counts and those specifically mentioned in multiples, the total count comes to 8.\\
\\
Step 6:\\
Reading through the sentence again to ensure no vegetables are missed. The count remains at 8.\\

Step 7:\\
Write down the total count as a string in the format: {"answer": "number"} where 'number' is your calculated total count of vegetables.\\
\\
So, after following the above steps, the output will be {"answer": "8"}.\\\\}\\
\hline
\tabincell{p{18cm}}{\\Test Score on Llama-2-70B-chat: \textcolor{red}{0.885}\\\\}
\\
\hline
\end{tabular}
}
\caption{Best prompt obtained with Automatic Behavior Optimization on Llama-2-70B-chat, object\_counting.}
\label{tab:ABO_best_llama1}
\end{table*}

\begin{table*}[h]
\centering
\small
\scalebox{0.81}{
\begin{tabular}{|c|}
\hline
\tabincell{p{18cm}}{\\1. Begin the task at the point '0'.\\
2. Read the instructions one by one and interpret them as follows:\\
   a. "Turn right/ left" changes your direction. It doesn't change your position but does affect the direction you'll move next. To simplify, assign "+" for right and "-" for left.\\
   b. "Always face forward" or "Turn around" changes the direction you're facing. It doesn't alter your position but changes the way you'll move. When you "turn around", it means you switch your direction by 180 degrees.\\
   c. "Take X steps" updates your position. If you're moving in the direction you're facing (forward), add X to your position. If you're moving in the opposite direction (backward), subtract X from your position.\\
   d. "Take X steps left/ right" asks you to move left or right relative to your current direction. If you're facing 'forward', assign "+" for right steps and "-" for left steps, and vice versa if facing 'backward'.\\ 
   e. "Take steps forward/ backward" depends on your current facing direction. 'Forward steps' add to your position if facing 'forward' and subtract if facing 'backward'. Similarly, 'backward steps' subtract from your position if facing 'forward' and add if facing 'backward'.\\
3. After interpreting all the instructions, if you're back at '0', return \{"answer": "A"\}. If you're at a position other than '0', return \{"answer": "B"\}.\\
\\
\#\#\#\# Examples\\
Here are some examples to help you thinking. \\
\\
\#\# Input\\
Take 3 steps. Take 5 steps. Turn around. Take 6 steps. Take 2 steps.\\
Choices:\\
A. True\\
B. False\\
\#\# Output\\
\\
To solve this task, I will follow the instructions given in the input one by one, and I will keep track of my direction and position. Initially, my direction is 'forward', because according to the instruction I always start facing forward, and my position is '0' because that's the starting point. \\
\\
I will interpret the instructions step by step:\\
\\
1. "Take 3 steps": This means to add 3 to my position because I'm moving forward. So my position changes from '0' to '3'.\\
\\
2. "Take 5 steps": Since I'm still facing forward, I add 5 to my position, so it changes from '3' to '8'.\\
\\
3. "Turn around": This means to change my direction by 180 degrees. Since I'm currently facing 'forward', I will now be facing 'backward'.\\
\\
4. "Take 6 steps": Since I'm now facing backward, taking a step will subtract from my position. Hence, I subtract 6 from my position, changing it from '8' to '2'.\\
\\
5. "Take 2 steps": I'm still facing backward, so I subtract 2 from my position. This changes my position from '2' to '0'.\\
\\
Now, since I am back at the '0' position, according to the task rule, I will return \{"answer": "A"\}.\\
\\
So, the answer is \{"answer": "A"\}.\\\\}\\
\hline
\tabincell{p{18cm}}{\\Test Score on Llama-2-70B-chat: \textcolor{red}{0.89}\\\\}
\\
\hline
\end{tabular}
}
\caption{Best prompt obtained with Automatic Behavior Optimization on Llama-2-70B-chat, navigate.}
\label{tab:ABO_best_llama2}
\end{table*}

\begin{table*}[h]
\centering
\small
\scalebox{0.81}{
\begin{tabular}{|c|}
\hline
\tabincell{p{18cm}}{\\Step 1: Carefully read both sentences.\\
\\
   Step 2: Identify the sentence that includes a degree of absurdity, exaggeration, or an unlikely scenario. This could potentially be the sarcastic statement. \\
\\
   Step 3: Evaluate the reality or truth of the sentences. However, remember that a statement being true does not necessarily mean it can't be sarcastic.\\
\\
   Step 4: Consider the context in which the statements are made, as context often plays a crucial role in understanding sarcasm. If the context is not provided or clear, focus on identifying any absurdity or improbability in the sentence.\\
\\
   Finally, provide the answer in the following format: \{"answer": "A"\} or \{"answer": "B"\} depending on which sentence is sarcastic.\\
\\
\#\#\#\# Examples\\
Here are some examples to help you thinking. \\
\\
\#\# Input\\
(a) Well, Alabama did beat the 1st ranked team. Surely that should be enough to get them in. (b) Well, Alabama did beat the 17th ranked team. Surely that should be enough to get them in.\\
Choices:\\
A. (a)\\
B. (b)\\
\#\# Output\\
\\
Step 1: Carefully read both sentences.\\
\\
Sentence (a): "Well, Alabama did beat the 1st ranked team. Surely that should be enough to get them in."
Sentence (b): "Well, Alabama did beat the 17th ranked team. Surely that should be enough to get them in."\\
\\
Step 2: Identify the sentence that includes a degree of absurdity, exaggeration, or an unlikely scenario.\\ 
\\
Sentence (a) is a statement of triumph while sentence (b) is a bit of an understatement. Beating the 17th ranked team, while commendable, is less significant than beating the 1st ranked team.\\
\\
Step 3: Evaluate the reality or truth of the sentences.\\
\\
Both statements could be true, depending on the actual outcomes of the games. \\
\\
Step 4: Consider the context in which the statements are made. \\
\\
Without context regarding the specific games or the overall standings, it's hard to definitively identify the sarcasm. However, the exaggeration or absurdity suggested by the idea that beating the 17th ranked team should guarantee a place in the tournament leads towards sentence (b) being the sarcastic statement.\\
\\
After taking into account the whole process, the answer is:\\
\\
\{"answer": "B"\}\\\\}\\
\hline
\tabincell{p{18cm}}{\\Test Score on Llama-2-70B-chat: \textcolor{red}{0.811}\\\\}
\\
\hline
\end{tabular}
}
\caption{Best prompt obtained with Automatic Behavior Optimization on Llama-2-70B-chat, snarks.}
\label{tab:ABO_best_llama3}
\end{table*}

\begin{table*}[h]
\centering
\small
\scalebox{0.81}{
\begin{tabular}{|c|}
\hline
\tabincell{p{18cm}}{\\Your task revolves around selecting the most appropriate question out of the provided options based on a provided short answer and its context. \\
\\
1. First, study the given short answer carefully. \\
\\
2. Review each of the provided question choices in relation to the short answer. Look for the question that the short answer wholly answers, not just partially. Remember, the short answer should be the precise and full answer to the selected question.\\
\\
3. Then, use the provided context to support your choice. The context should resonate with the selected question and the short answer but should not be the primary determinant of the chosen question. \\
\\
4. Once you've chosen the most appropriate question, format your response as a JSON object. This should have the key "answer" and the value should be the letter ("A", "B", "C", etc.) corresponding to the right question. \\
\\
For instance, if the correct question is the first option, your answer should be \{"answer": "A"\}.\\
\\
Note: While the context might relate to all choices, the correct selection should be the one that the short answer fully and directly responds to. Do not rely solely on the context. The exactness of the correlation between the short answer and the question is critical.\\
\\
\#\#\#\# Examples\\
Here are some examples to help you thinking. \\
\\
\#\# Input\\
Short Answer: twice \\
Context: Oxygen is more soluble in water than nitrogen is. Water in equilibrium with air contains approximately 1 molecule of dissolved O
2 for every 2 molecules of N2, compared to an atmospheric ratio of approximately 1:4. The solubility of oxygen in water is temperature-dependent, and about twice as much (14.6 mg·L-1) dissolves at 0 °C than at 20 °C (7.6 mg·L-1). At 25 °C and 1 standard atmosphere (101.3 kPa) of air, freshwater contains about 6.04 milliliters (mL) of oxygen per liter, whereas seawater contains about 4.95 mL per liter. At 5 °C the solubility increases to 9.0 mL (50\% more than at 25 °C) per liter for water and 7.2 mL (45\% more) per liter for sea water.\\
Choices:\\
A. At 0 degrees celsius, how much more oxygen dissolves than at 20 degrees celsius?\\
B. Under normal circumstances, how much oxygen can be contained in a liter of fresh water?\\
\#\# Output\\
\\
First, I'll look at the short answer: "twice". This suggests that the right question will ask about a ratio, comparison, or frequency, and the answer to that question will be twice of something.\\
\\
Next, I'll consider each question in light of the short answer:\\
\\
A. At 0 degrees celsius, how much more oxygen dissolves than at 20 degrees celsius?\\
B. Under normal circumstances, how much oxygen can be contained in a liter of fresh water?\\
\\
Now, I'll check for the exactness of the correlation between the short answer and each question:\\
\\
Question A is asking for a comparison of how much more oxygen dissolves at 0 degrees celsius than at 20 degrees celsius. In the context, it's stated that "twice as much (14.6 mg·L-1) dissolves at 0 °C than at 20 °C (7.6 mg·L-1)". Hence, the short answer "twice" fully and directly answers this question.\\
\\
On the other hand, Question B is asking for a specific amount of oxygen contained in a liter of fresh water under normal circumstances. The answer to this question would require a specific numeric value, not a comparative measure like "twice". The context provides this value as "6.04 milliliters (mL) of oxygen per liter" at 25 °C and 1 standard atmosphere of air, but it doesn't match the short answer "twice".\\
\\
Hence, based on the short answer and analyzing the context, it's clear that the most fitting question is A.\\
\\
So, the output is \{"answer": "A"\}.\\\\}\\
\hline
\tabincell{p{18cm}}{\\Test Score on Llama-2-70B-chat: \textcolor{red}{0.81}\\\\}
\\
\hline
\end{tabular}
}
\caption{Best prompt obtained with Automatic Behavior Optimization on Llama-2-70B-chat, question\_selection.}
\label{tab:ABO_best_llama4}
\end{table*}

\begin{table*}[h]
\centering
\small
\scalebox{0.81}{
\begin{tabular}{|c|}
\hline
\tabincell{p{18cm}}{\\1. Begin by reading the sentence very carefully, understanding all the objects and the question at the end.\\
   \\
   2. Identify all objects mentioned in the sentence. \\
\\
   3. Determine the context of the question, that is, what type of objects the question is asking for.\\
\\
   4. Create two lists: one for the objects that fall within the context of the question, and another for those that do not.\\
\\
   5. Begin counting the objects that fall within the context of the question, note the number associated with each. If no number is specified, assume it to be one. For example, 'a lamp' should be counted as one lamp.\\
\\
   6. Verify that the objects being counted are indeed relevant to the context of the question.\\
\\
   7. Start a running total of all the relevant objects.\\
\\
   8. Compare each object in the list of irrelevant objects with the context of the question. If it matches, move it to the list of relevant objects and adjust the total count accordingly. \\
\\
   9. Format the total count into a string.\\
\\
   10. Create a JSON object with the key as "answer" and set the total count string as its value. \\
\\
   11. Return this JSON object.\\
\\
\#\#\#\# Examples\\
Here are some examples to help you thinking. DONOT simply copy the outputs in the examples!\\
\\
\#\# Input\\
I have a grape, and two bananas. How many fruits do I have?\\
\#\# Output\\
\\
1. Reading the sentence: "I have a grape, and two bananas. How many fruits do I have?"\\
   \\
2. Identifying all objects mentioned in the sentence: grape, bananas. \\
\\
3. Determining the context of the question: The question is asking for the total number of fruits. \\
\\
4. Creating two lists: one for the objects that fall within the context of the question, and another for those that do not. \\
   - Relevant objects: grape, bananas. \\
   - Irrelevant objects: None.\\
\\
5. Counting the objects that fall within the context of the question. \\
   - grape: 1 \\
   - bananas: 2\\
\\
6. Verification: The objects being counted (grape and bananas) are indeed relevant to the context of the question (fruits).\\
\\
7. Running total of all the relevant objects: 1 (grape) + 2 (bananas) = 3\\
\\
8. Comparing each object in the list of irrelevant objects with the context of the question. There are no irrelevant objects in this context.\\
\\
9. Formatting the total count (3) into a string: "3"\\
\\
10. Creating a JSON object with the key as "answer" and set the total count string as its value: \{"answer": "3"\}\\
\\
11. Returning the JSON object: \{"answer": "3"\}\\\\}\\
\hline
\tabincell{p{18cm}}{\\Test Score on GPT-3.5-Turbo: \textcolor{red}{0.975}\\\\}
\\
\hline
\end{tabular}
}
\caption{Best prompt obtained with Automatic Behavior Optimization on GPT-3.5-Turbo, object\_counting.}
\label{tab:ABO_best_35_1}
\end{table*}

\begin{table*}[h]
\centering
\small
\scalebox{0.81}{
\begin{tabular}{|c|}
\hline
\tabincell{p{18cm}}{\\a. Begin by understanding that the starting point is always the origin, represented by the coordinates (0,0).\\
\\
b. Break down the navigation instructions one by one.\\
\\
c. Define a concept of current direction. Start with the assumption that the direction is 'up'.\\
\\
d. For each instruction: \\
   - If the instruction is to take a number of steps "forward", determine the change in coordinates based on the current direction. If the current direction is 'up', increase the y-coordinate by the amount of steps. If 'down', decrease the y-coordinate. If 'right', increase the x-coordinate. If 'left', decrease the x-coordinate.\\
   - If the instruction is to take a number of steps "backward", determine the change in coordinates opposite to the current direction. If the current direction is 'up', decrease the y-coordinate by the amount of steps. If 'down', increase the y-coordinate. If 'right', decrease the x-coordinate. If 'left', increase the x-coordinate.\\
   - If the instruction is to turn right, update the current direction to the right of the current direction.\\
   - If the instruction is to turn left, update the current direction to the left of current direction.\\
   - If the instruction is to turn around, reverse the current direction.\\
\\
e. After each movement, update your position based on the instruction and keep track of these changes.\\
\\
f. After all the instructions are processed, compare your ending point to the starting point, which is the origin (0,0).\\
\\
g. If the ending point is (0,0), then the answer is {"answer": "A"}, indicating the instructions lead back to the starting point.\\
\\
h. If the ending point is not (0,0), then the answer is {"answer": "B"}, indicating the instructions do not lead back to the starting point.\\
\\
\#\#\#\# Examples\\
Here are some examples to help you thinking. DONOT simply copy the outputs in the examples!\\
\\
\#\# Input\\
Always face forward. Take 10 steps left. Take 10 steps forward. Take 7 steps forward. Take 2 steps forward.\\
Choices:\\
A. True\\
B. False\\
\#\# Output\\
\\
To begin the process of understanding the given instructions, let's interpret them while keeping the initial position at the origin (0,0) in focus. The first direction gives us an idea of our starting orientation, which is 'forward'. As per direction 'c', we'll assume that 'forward' equates to moving 'up', so the current direction is 'up'. \\
\\
Now, let's process each instruction one by one:\\
\\
1. "Always face forward." - This doesn't give us a direction of movement, so the coordinates remain (0,0) and the current direction stays 'up'.\\
2. "Take 10 steps left." - In this case, considering our current direction is 'up', 'left' would be decreasing x-coordinate. So, we'll decrease the x-coordinate by 10 steps, which changes our coordinates from (0,0) to (-10,0). The current direction remains 'up'.\\
3. "Take 10 steps forward." - Considering the current direction is 'up', 'forward' would mean increasing the y-coordinate. So, the y-coordinate is increased by 10 steps, changing the coordinates from (-10,0) to (-10,10).\\
4. "Take 7 steps forward." - Again, given the current direction is 'up', 'forward' correlates to an increase in the y-coordinate. So, increase the y-coordinate by 7 steps, changing the coordinates from (-10,10) to (-10,17).\\
5. "Take 2 steps forward." - Similarly, with the current direction being 'up', 'forward' means an increase in y-coordinate. Increase the y-coordinate by 2 steps, which changes our coordinates from (-10,17) to (-10,19).\\
\\
After following all the instructions, we find the final position to be (-10,19). This is not the same as the origin (0,0), the starting point.\\
\\
Following the last two points in the instructions (g and h), since the ending position is not the same as the origin, the answer is {"answer": "B"}. This indicates that the instructions do not lead back to the starting point. \\
\\
\#\# Output\\
{"answer": "B"}\\\\}\\
\hline
\tabincell{p{18cm}}{\\Test Score on GPT-3.5-Turbo: \textcolor{red}{0.985}\\\\}
\\
\hline
\end{tabular}
}
\caption{Best prompt obtained with Automatic Behavior Optimization on GPT-3.5-Turbo, navigate.}
\label{tab:ABO_best_35_2}
\end{table*}

\begin{table*}[h]
\centering
\small
\scalebox{0.81}{
\begin{tabular}{|c|}
\hline
\tabincell{p{18cm}}{\\1. Read both sentences (a) and (b) thoroughly.\\
2. Try to understand the general sentiment or expectation surrounding the situation or subject in real life. This is because sarcasm often works by stating the opposite of what is generally accepted or expected.\\
3. Analyze each sentence for signs of irony by:\\
    i. Identifying the literal meaning of the sentence.\\
    ii. Comparing this literal meaning to the expected perception or sentiment.\\
    iii. Looking for contradictions between the literal meaning and expected perception - this indicates irony.\\
4. Look for any exaggerations or understatements. Sarcasm often involves either grossly over-exaggerating or downplaying a situation.\\
5. Based on your analysis, decide which sentence is sarcastic. Remember that it could be either, neither, or both, depending on the context.\\
6. Provide your answer in the format: {"answer": "A"} or {"answer": "B"}, where "A" corresponds to sentence (a) and "B" corresponds to sentence (b).\\
\\
\#\#\#\# Examples\\
Here are some examples to help you thinking. DONOT simply copy the outputs in the examples!\\
\\
\#\# Input\\
(a) And the likelihood of internet-connected cars being hacked and controlled remotely,....zero. \\
(b) And the likelihood of internet-connected cars being hacked and controlled remotely,....high.\\
Choices:\\
A. (a)\\
B. (b)\\
\#\# Output\\
\\
1. Reading both sentences thoroughly:\\
    - Sentence (a): "And the likelihood of internet-connected cars being hacked and controlled remotely,....zero."\\
    - Sentence (b): "And the likelihood of internet-connected cars being hacked and controlled remotely,....high."\\
\\
2. Understanding the general sentiment or expectation:\\ 
   - In the real world, as technology advances and everything gets connected to the internet (including cars), the chances of them being hacked and controlled remotely increases due to the vulnerabilities that might arise from such connections. \\
\\
3. Analyzing each sentence for signs of irony:\\
\\
    - Sentence (a):\\
          i. The literal meaning of the sentence is that there is absolutely no chance of internet-connected cars being hacked and controlled remotely.\\
          ii. This contradicts the general perception that connecting cars to the internet could increase the chances of them being hacked.\\
          iii. There is a clear contradiction between the literal meaning of the sentence and the expected perception, which could suggest irony.\\
\\
    - Sentence (b):\\
          i. The literal meaning of the sentence is that there's a high chance of internet-connected cars being hacked and controlled remotely.\\
          ii. This aligns with the general sentiment that connecting cars to the internet could increase their vulnerability to hacking.\\
          iii. In this case, there is no contradiction between the literal meaning and the expected perception, so no signs of irony.\\
\\
\\
4. Looking for exaggerations or understatements:
    - In sentence (a), stating that the likelihood of internet-connected cars being hacked is "zero" could be seen as an understatement, given the general perception of the subject. On the other hand, sentence (b) seems to align with general sentiments and doesn't exaggerate or understate the situation.\\
\\
5. Based on the analysis, sentence (a) is likely to be sarcastic due to the contradiction between the literal meaning and the real-world perception of the topic, as well as the potential understatement involved in stating the hacking likelihood as "zero".\\
\\
6. Therefore, the answer is:\\
\\
\{"answer": "A"\}\\\\}\\
\hline
\tabincell{p{18cm}}{\\Test Score on GPT-3.5-Turbo: \textcolor{red}{0.802}\\\\}
\\
\hline
\end{tabular}
}
\caption{Best prompt obtained with Automatic Behavior Optimization on GPT-3.5-Turbo, snarks.}
\label{tab:ABO_best_35_3}
\end{table*}

\begin{table*}[h]
\centering
\small
\scalebox{0.81}{
\begin{tabular}{|c|}
\hline
\tabincell{p{18cm}}{\\1. Start by reading the short answer and the context paragraph carefully. Try to understand the main subject as well as any specific details or associations mentioned.\\
\\
2. Identify and highlight any keywords in the short answer. These keywords are what you will look for in the context paragraph and the question choices.\\
\\
3. Review the question choices. Make sure you understand what each question is asking for. \\
\\
4. Match the identified keywords from the short answer to the contents of the question choices. However, remember to not just match the keywords, but also consider the detailed context in which these are mentioned in both the short answer and question choices.\\
\\
5. Once you believe you have the matched question, go back to the context paragraph. Verify that the relationship or association implied by your selected question matches the one found in the context paragraph.\\
\\
6. If everything aligns, format your response as a JSON object with the key as "answer" and the value as the letter corresponding to your chosen question. Ensure to double-check your answer before finalizing it.\\
\\
\#\#\#\# Examples\\
Here are some examples to help you thinking. DONOT simply copy the outputs in the examples!\\
\\
\#\# Input\\
Short Answer: 1830 \\
Context: France took control of Algeria in 1830 but began in earnest to rebuild its worldwide empire after 1850, concentrating chiefly in North and West Africa, as well as South-East Asia, with other conquests in Central and East Africa, as well as the South Pacific. Republicans, at first hostile to empire, only became supportive when Germany started to build her own colonial empire. As it developed, the new empire took on roles of trade with France, supplying raw materials and purchasing manufactured items, as well as lending prestige to the motherland and spreading French civilization and language as well as Catholicism. It also provided crucial manpower in both World Wars.\\
Choices:\\
A. When did french republicans begin to support the expansion of the french empire?\\
B. When did france begin to restore its global empire in earnest?\\
C. When did france seize power in algeria?\\
\#\# Output\\
\\
The short answer provides the keyword "1830". Reading the context paragraph, "1830" is associated with the time when "France took control of Algeria". This keyword and context aligns with the question choices C "When did france seize power in algeria?" as it directly asks for the year in which France gained control over Algeria, which is the information provided in the short answer.\\
\\
The other questions do not align with the short answer. Question A is about the time when French republicans started supporting the empire's expansion, which is not discussed in the short answer. Question B is about when France began to earnestly restore its global empire, which according to the context paragraph, happened after 1850, not 1830.\\
\\
Going back to the context paragraph, the relationship implied by the chosen question - that France took control of Algeria in 1830 - matches the one provided in the context.\\
\\
So the matched question is C. Formatting the response as a JSON object:\\
\\
{"answer": "C"}\\\\}\\
\hline
\tabincell{p{18cm}}{\\Test Score on GPT-3.5-Turbo: \textcolor{red}{0.905}\\\\}
\\
\hline
\end{tabular}
}
\caption{Best prompt obtained with Automatic Behavior Optimization on GPT-3.5-Turbo, question\_selection.}
\label{tab:ABO_best_35_4}
\end{table*}


\end{document}